%% file: main.tex
\documentclass[11pt]{article}
\usepackage[colormark, titleright, nolineno]{showlab}
\shorttitle{\textbf{\textcolor{perceive}{P}\textcolor{anticipate}{A}\textcolor{plan}{P}\textcolor{act}{A}\textcolor{verify}{V}}}

\logo{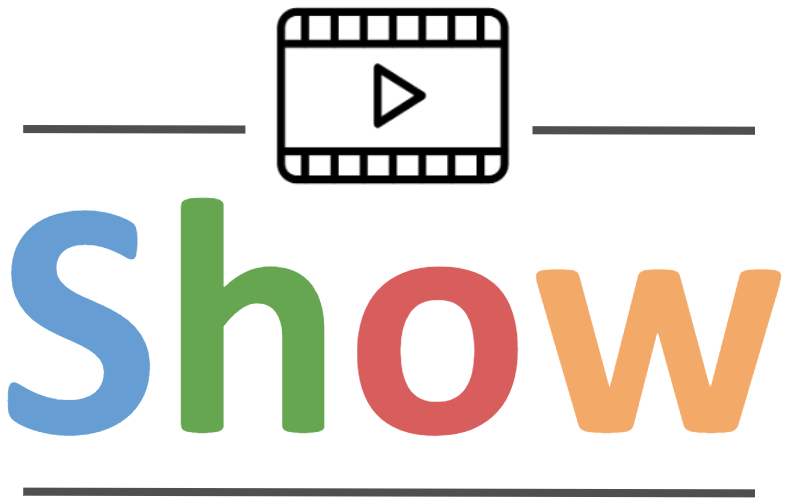}   % replace with a vector .pdf when available
\logoheight{16mm}       % size of the large logo, only used with [titlelogo]
\headlogoheight{6.5mm}  % size of the logo in the running head
\runningmark{October, 2026} % text at the top right of every page

\title{
Multimodal Embodied Agents: A Unified Capability-centric Perspective from Computer-Use to Robot-Use
}

\author{%
  Yanzhe Chen\amark{1}\quad
  Ziyi Yang \amark{1}\quad
  Jifeng Zhu \amark{1}\quad
  Qiming Huang\amark{1}\quad
  Ruihe An \amark{1}\quad
  Peiyao Xu \amark{1}\quad \\
  Hesen Yang \amark{1}\quad
  Runda Liu \amark{1}\quad
  Chang Gong \amark{1}\quad
  Zhijun Cao\amark{1}\quad
  Zechen Bai\amark{1}\quad
  Wenzheng Zeng\amark{1}\quad  \\
  Yiqi Lin\amark{1}\quad
  Guoqiang Liang\amark{1}\quad
  Kevin Yuchen Ma\quad
  Kevin Qinghong Lin\quad
  Mike Zheng Shou\amark[\textdagger]{1}%
}
\affiliation{Show Lab, National University of Singapore}
\authornote{\textsuperscript{\textdagger}Corresponding author}
\links{%
  \resource{\textbf{Project Page}}{https://showlab.github.io/Awesome-Multimodal-Embodied-Agent}%
}
\date{}

\definecolor{qimingGreen}{HTML}{1B5E20}

\usepackage{longtable}
\usepackage{booktabs}
\usepackage{array}
\usepackage{caption}
\usepackage{wasysym}

\usepackage{graphicx} % 提供 \resizebox
\usepackage{forest}   % 编译 taxonomy.tex
\usepackage[numbers,sort&compress]{natbib}

\usepackage{tabularx}
\usepackage{makecell}

\newcommand{\directcov}{\CIRCLE}
\newcommand{\partialcov}{\LEFTcircle}
\newcommand{\nonecov}{\Circle}
\newcolumntype{L}[1]{>{\raggedright\arraybackslash}p{#1}}
\newcolumntype{C}[1]{>{\centering\arraybackslash}p{#1}}

\usepackage{amssymb} % provides \checkmark
\newcommand{\surveyfull}{%
  \textcolor{qimingGreen}{\ensuremath{\checkmark}}}
\newcommand{\surveynone}{%
  \textcolor{orange}{\ensuremath{\times}}}

\usepackage[table]{xcolor}
\definecolor{mmembodied}{HTML}{CFEEDC}
\definecolor{mmagent}{HTML}{B7F0F0}
\definecolor{robotics}{HTML}{E7ECC8}

\definecolor{subcategory}{HTML}{F6F6F6}
\definecolor{tabledivider}{HTML}{E8E8E8}
\usepackage{adjustbox}
\usepackage{makecell}
\definecolor{perceive}{HTML}{F57C6E}
\definecolor{anticipate}{HTML}{F2B56F}
\definecolor{plan}{HTML}{84C3B7}
\definecolor{act}{HTML}{71B7ED}
\definecolor{verify}{HTML}{B8AEEB}
\definecolor{lightgreen}{RGB}{240, 251, 237} % 

\providecommand{\papav}{%
  \textcolor{perceive}{P}%
  \textcolor{anticipate}{A}%
  \textcolor{plan}{P}%
  \textcolor{act}{A}%
  \textcolor{verify}{V}%
}
\usepackage{float}
\input{sections/1-introduction/evo_tree/clickable_tree.tex}
\begin{document}
\maketitle
\thispagestyle{firstpage}
\input{sections/0-abstract}

% \keywords{multimodal agents, vision-language-action models, embodied AI, world models, evaluation}
% ---------------------------------------------------------------------

\input{sections/1-introduction/introduction}
\input{sections/2-background/background}
\input{sections/3-capability}
\input{sections/4-benchmark}

\input{sections/5-challenges}
\input{sections/6-conclusions}

% ---------------------------------------------------------------------
\bibliographystyle{plainnat}   % numeric [1]; swap to unsrtnat for order-of-appearance
\bibliography{refs}

\end{document}

%% file: sections/0-abstract.tex
\begin{abstract}

% 起
Multimodal agents (MMAs) sustain interaction through reasoning, memory, tools, and feedback, most visibly as computer-use agents, while robotic systems (RSs) couple sensing and actuation under physical dynamics. We define \emph{multimodal embodied agents} (MMEAs) as goal-directed systems that couple multimodal task reasoning with physical action and revise decisions from the resulting feedback. This raises a central question: \textit{what changes when a multimodal agent moves from computer-use to robot-use?}
% 承
We introduce \textbf{\papav{}}, a unified capability-centric framework of five recurring functions:
\textcolor{perceive}{\textbf{P}}erceive the current state,
\textcolor{anticipate}{\textbf{A}}nticipate action effects, \textcolor{plan}{\textbf{P}}lan a feasible course, \textcolor{act}{\textbf{A}}ct through an interface or body, and \textcolor{verify}{\textbf{V}}erify the outcome.
% Why
Defined by function rather than architecture, these capabilities let PAPAV identify what all three share and then compare MMEAs with MMAs and with RSs.
% 转
Five physical constraints, from partial observability to unverifiable outcomes, leave fewer choices fixed in advance and less room to reverse errors.
Across 64 benchmarks, explicit evaluation centers on Act; only 5 assess Anticipate and 3 assess Verify, leaving both largely hidden behind task success.
% 合
These constraints also frame the open challenges we identify. PAPAV therefore provides a common basis for designing reliable physical agents and measuring progress across all five capabilities rather than task success alone.

% Multimodal agents (MMAs) sustain interaction through reasoning, memory, tools, and feedback, most visibly as computer-use agents, while robotic systems (RSs) couple sensing and actuation under physical dynamics. We define multimodal embodied agents (MMEAs) as goal-directed systems that couple multimodal task reasoning with physical action and revise decisions from the resulting feedback. This raises a central question: what changes when a multimodal agent moves from computer-use to robot-use? We introduce PAPAV, a unified capability-centric framework of five recurring functions: Perceive the current state, Anticipate action effects, Plan a feasible course, Act through an interface or body, and Verify the outcome. Defined by function rather than architecture, these capabilities let PAPAV identify what all three share and then compare MMEAs with MMAs and with RSs. Five physical constraints, from partial observability to unverifiable outcomes, leave fewer choices fixed in advance and less room to reverse errors. Across 64 benchmarks, explicit evaluation centers on Act; only 5 assess Anticipate and 3 assess Verify, leaving both largely hidden behind task success. These constraints also frame the open challenges we identify. PAPAV therefore provides a common basis for designing reliable physical agents and measuring progress across all five capabilities rather than task success alone.

\end{abstract}

%% file: sections/1-introduction/introduction.tex
\begin{figure}[H]
  \centering
  \includegraphics[width=\linewidth]{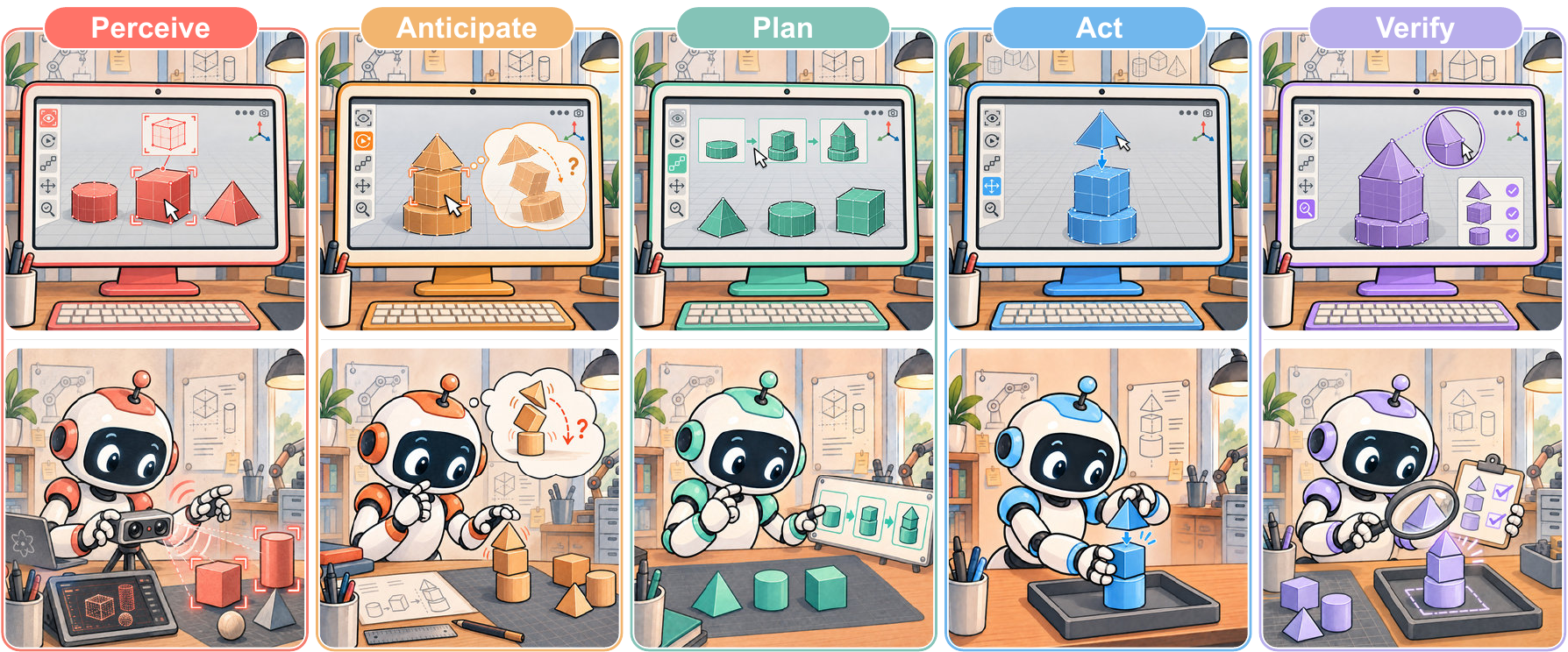}
  \caption{\textbf{A Unified Capability-centric Perspective (PAPAV) from computer-use to robot-use.}
  \textcolor{perceive}{\textbf{Perceive}},
  \textcolor{anticipate}{\textbf{Anticipate}},
  \textcolor{plan}{\textbf{Plan}},
  \textcolor{act}{\textbf{Act}}, and
  \textcolor{verify}{\textbf{Verify}}
  are five functional roles that recur in goal-directed interaction, whether an agent operates a computer (top) or a robot (bottom).
  Defined by function rather than architecture, they place both settings on a common coordinate and isolate what changes once actions acquire physical consequences.
  }
  \label{fig:papav-overview}
\end{figure}

% \clearpage
\section{Introduction}
\label{sec:intro}

Multimodal agents (MMAs) have grown from one-shot generation into systems that interact over time, use external tools, and revise decisions from feedback, and they now operate browsers, desktops, and phones as computer-use agents~\citep{yao2023react,yang2023mm,shen2023hugginggpt,shinn2023reflexion,wang2024voyager,zhou2024webarena,xie2024osworld}. Robotic systems (RSs) have followed a complementary path: foundation models and generalist policies increasingly connect language and perception to physical action across a widening range of tasks~\citep{ahn2022can,huang2022inner,liang2023code,driess2023palm,brohan2023rt1,brohan2023rt2,jiang2023vima,oxe2024,octo2024,black2024pi0}. Robotics offers a mature account of how to \emph{close the loop} under physical dynamics~\citep{hogan1985impedance,kaelbling2011hierarchical,finn2017deep,ebert2017self,kahn2018composable,campos2021orb,garrett2021tamp}, while agent research offers ways to \emph{sustain the loop} through longer interactions~\citep{yao2023react,shinn2023reflexion,wang2024voyager,koh2024visualwebarena,chen2025code2video,he2024webvoyager, chen2026showharness, chen2026eta}.

We use \emph{multimodal embodied agent} (MMEA) for a goal-directed system that uses multimodal evidence to guide physical action and feedback from that action to revise later decisions. Bringing the two traditions together changes more than the action interface. Computer-use agents can draw on structured state and direct feedback supplied by software, while conventional robotic systems can rely on constraints chosen for a known operating setting~\citep{zhou2024webarena,xie2024osworld,yao2024taubench,campos2021orb,garrett2021tamp}. Open-ended physical interaction offers neither in full: the same loop becomes less predictable and less forgiving, and more decisions move into the interaction itself. This motivates our central question:
\textbf{\emph{What changes when a multimodal agent moves from computer-use to robot-use?}}

  \begin{figure}[t]
    \centering
    \renewcommand{\TreeImageFile}{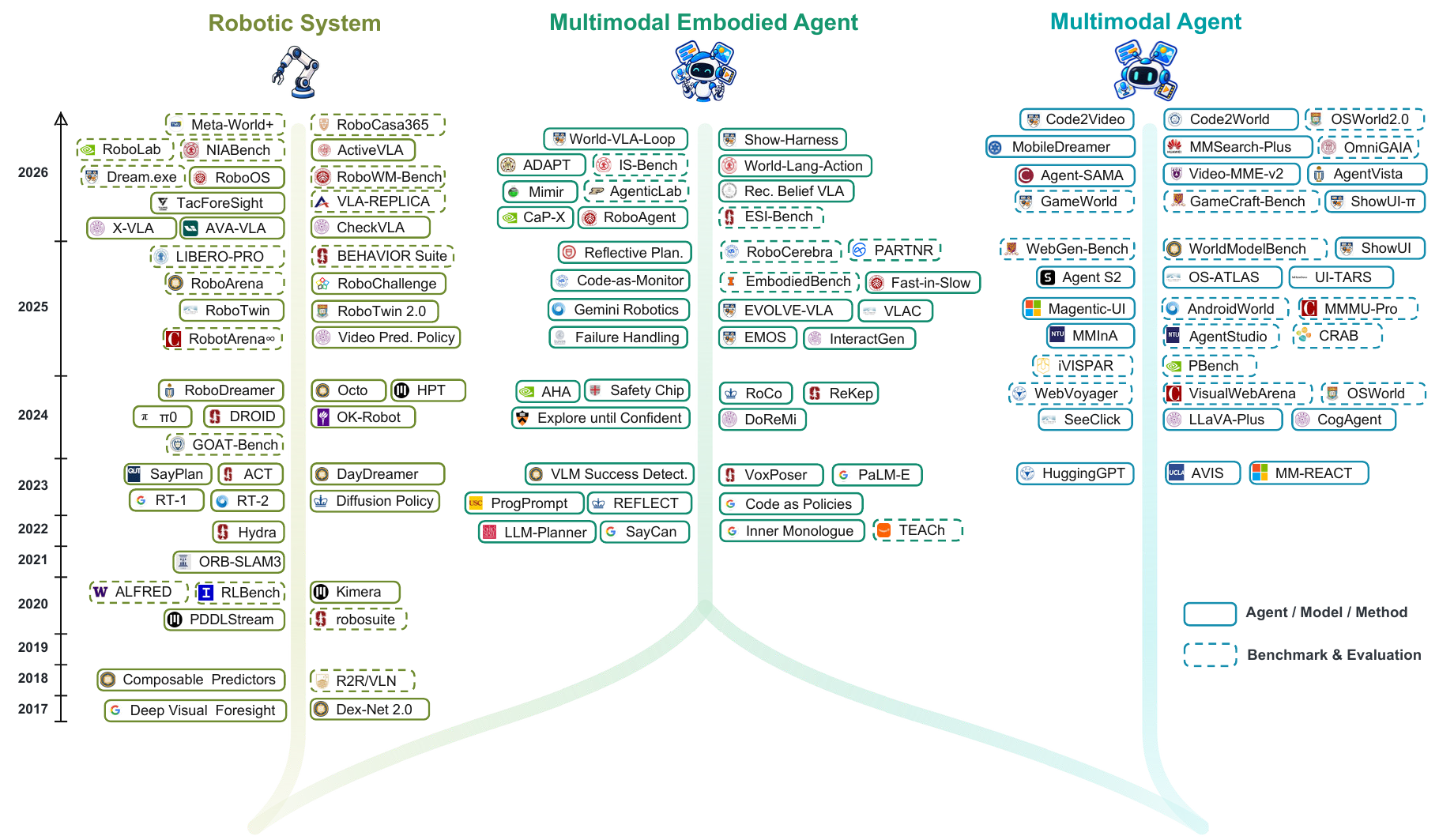}
    \renewcommand{\TreeFigureWidth}{\linewidth}
    \ClickableTreeGraphic
    \caption{%
      \textbf{Chronological overview of representative work.}
      Robotic systems progress from visual foresight and task planning to generalist policies, world models, and multi-site evaluation; multimodal agents from tool-augmented reasoning to GUI grounding and computer-use benchmarks.
      The middle trunk appears in 2022 as language-model planners meet robot skills and keeps widening.
    }
    \label{fig:evolution-tree}
  \end{figure}

\paragraph{A unified basis for cross-domain comparison}
Survey taxonomies still tend to follow the field in which they were developed. Surveys of computer-use agents foreground computational architectures and digital interaction~\citep{durante2024agentaisurveyinghorizons,xie2024largemultimodalagentssurvey,tang2025surveymllmbasedguiagents,yao2025surveyagenticmultimodallarge,v2026agenticartificialintelligenceai}, whereas surveys of embodied AI, VLA models, and robot autonomy emphasize physical tasks and execution~\citep{liu2025aligningcyberspacephysical,ma2026surveyvisionlanguageactionmodelsembodied,salimpour2025embodiedagenticaireview,tu2026large}. Several recent accounts cover both computer-use and robot-use~\citep{durante2024agentaisurveyinghorizons,liu2025aligningcyberspacephysical,salimpour2025embodiedagenticaireview,yao2025surveyagenticmultimodallarge,v2026agenticartificialintelligenceai,mokaria2026survey}, but covering both is different from drawing the analogy between them (Table~\ref{tab:survey-position}): existing categories describe systems within each setting and do not show how the same function changes when its actions acquire physical consequences. The need for such an analogy is growing. Foundation models, world models, and broader robot datasets now connect capabilities that were once developed separately~\citep{driess2023palm,brohan2023rt2,bruce2024genie,hafner2025dreamerv3,yang2024unisim,oxe2024,khazatsky2024droid}, yet strong components do not by themselves form a coherent physical agent. Coherence depends on coordination across the full loop as the environment changes, which calls for a unit of comparison that remains stable across both settings. Figure~\ref{fig:evolution-tree} traces the two research paths and their recent convergence.

\input{sections/1-introduction/survey_table}
\paragraph{A capability view of the task loop}
We introduce \textbf{\papav{}}, a unified capability-centric perspective built around five functions in a recurring \emph{task loop}, the cycle through which an agent advances a task one step at a time. \textcolor{perceive}{\textbf{P}}erceive establishes the current state; \textcolor{anticipate}{\textbf{A}}nticipate estimates the effects of candidate actions; \textcolor{plan}{\textbf{P}}lan selects a feasible course; \textcolor{act}{\textbf{A}}ct realizes that course through an interface or body; and \textcolor{verify}{\textbf{V}}erify judges the outcome and whether further interaction is needed. Each function is defined by what it contributes to the loop rather than by the component that implements it: one model may support several functions, and one function may be distributed across several models or tools. A click, a skill call, and a robot trajectory differ in form, but PAPAV asks what each contributes to the same task loop rather than whether the systems share an architecture or action space. PAPAV also sharpens two boundaries often folded into a standard observation--decision--action loop: Anticipate asks what a candidate action may cause before Plan chooses a course, and Verify asks whether post-action evidence supports the intended outcome rather than merely recording another observation. These distinctions make prediction and outcome judgment visible as separate functions while remaining compatible with existing agent and control-loop taxonomies~\citep{yao2025surveyagenticmultimodallarge,liu2025aligningcyberspacephysical,salimpour2025embodiedagenticaireview,mokaria2026survey,tu2026large}.

\begin{figure}[!t]
  \centering
  \resizebox{\linewidth}{!}{%
    \input{figures/taxonomy/taxonomy.tex}%
  }
  \caption{\textbf{Research landscape and taxonomy of PAPAV.}
  The capability branch groups representative work by non-exclusive evidence lineages; the benchmark branch groups the curated corpus by primary evaluation target.}
  %  Neither placement determines whole-system MMEA membership, which is defined separately in Section~\ref{sec:papav-synthesis}.
  \label{fig:taxonomy}
\end{figure}
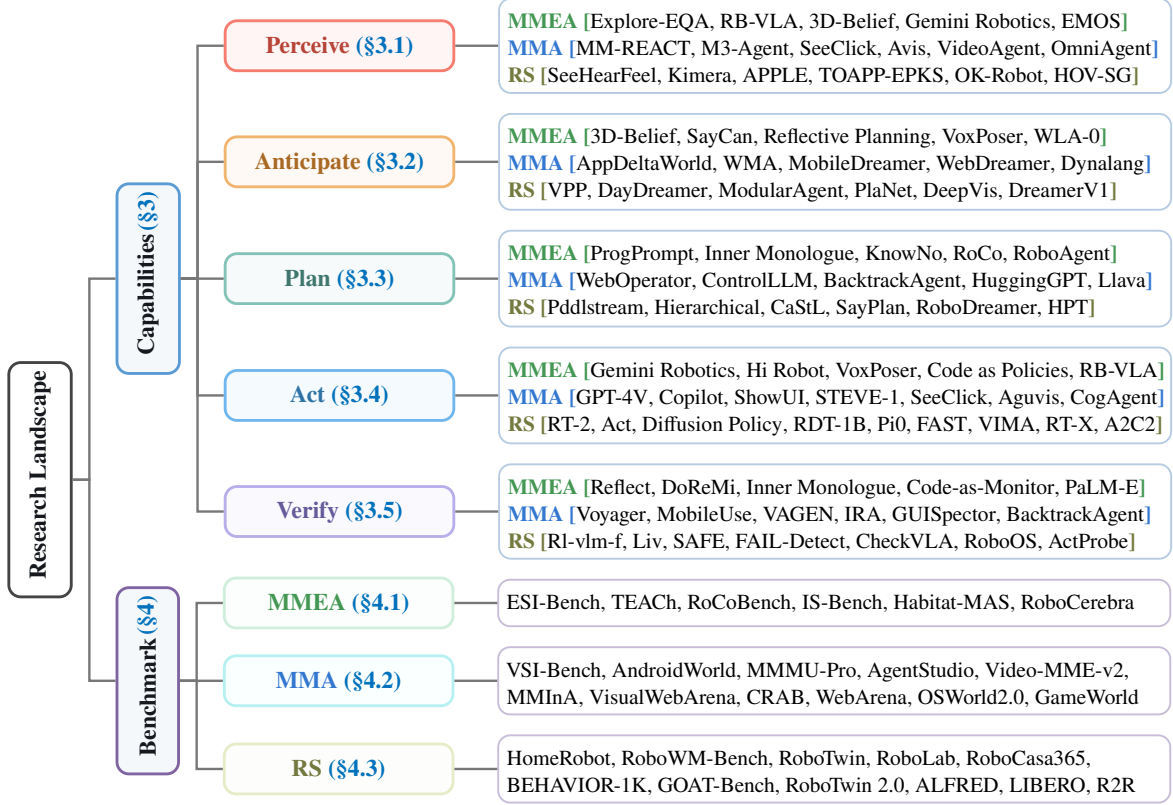

The shared coordinate then isolates what embodiment changes. We study five constraints that recur in physical interaction: \textbf{partial observability}, \textbf{limited simulatability}, \textbf{irreversibility}, \textbf{real-time coupling}, and \textbf{unverifiable outcomes}. Together they make the loop less predictable and less forgiving, increasing the weight of what the agent decides during interaction rather than what the interface or system design fixes in advance. Memory and orchestration keep the five functions connected across time; we call this coordinating substrate the \emph{agentic harness} rather than a sixth capability.

\paragraph{Position and scope of this survey} Our scope spans MMAs, conventional RSs, and their intersection. We borrow \emph{computer-use} and \emph{robot-use} as shorthand for the two ends of this transition~\citep{xie2024osworld,isola2026robotuse}, and use \emph{digital} and \emph{physical} throughout the body for the interaction settings along it. Robot-use here covers any configuration in which multimodal task reasoning acts through a robot body, by remote tool-style control or by an end-to-end policy, and is thus broader than cloud-controlled robots alone. The MMA and RS categories are system classes rather than settings, and they overlap. MMAs commonly operate through software interfaces, including games and simulated worlds~\citep{ouyang2026gameworld,wang2024voyager}, whereas RSs center body-specific sensing and control, even when evaluated in simulation. We include virtual embodied work when it informs a capability, while whole-system MMEA membership follows the physical task-loop criteria in Section~\ref{sec:bg-mea}. The task loop is therefore the unit of comparison: evidence for one capability does not by itself classify an entire system as an MMEA. Table~\ref{tab:survey-position} compares related surveys in terms of computer-use and robot-use coverage, benchmark analysis, and digital–physical analogy, highlighting PAPAV’s emphasis on a shared capability perspective across settings.

\paragraph{Contributions}
This survey makes three contributions:
\begin{itemize}

\item \textbf{A unified capability-centric perspective.} PAPAV enables unified comparison of MMAs, RSs, and MMEAs through five functional roles in a task loop, spanning computer-use and robot-use.
\item \textbf{A three-way capability comparison.} For each capability, we identify what MMAs, RSs, and MMEAs share, then separately analyze MMEA--MMA and MMEA--RS differences.
\item \textbf{A capability-level benchmark analysis.} We analyze PAPAV coverage in 64 benchmarks, distinguishing explicit capability evaluation from evidence available only through task outcomes. Explicit evaluation concentrates on Act, with few benchmarks isolating Anticipate or Verify.

\end{itemize}

Figure~\ref{fig:taxonomy} summarizes the research landscape: its capability branch organizes representative evidence by PAPAV relation and literature lineage, and its benchmark branch organizes the curated corpus by primary evaluation target. Neither placement is a shortcut for whole-system MMEA membership.

\paragraph{Organization}
Section~\ref{sec:background} defines the shared task-loop setting and PAPAV. Section~\ref{sec:capabilities} compares the five capabilities across digital and physical systems, with particular attention to the five physical constraints. Section~\ref{sec:benchmark} analyzes benchmark coverage at the capability level, Section~\ref{sec:challenges} discusses open challenges for reliable physical interaction, and Section~\ref{sec:conclusion} concludes.

%% file: sections/1-introduction/survey_table.tex
% Existing definitions:
% \surveyfull, \surveynone, \papav
% slTint, subcategory, lightgreen

\definecolor{tabledivider}{HTML}{E8E8E8}

\begin{table}[t]
\centering
\scriptsize
\setlength{\tabcolsep}{4pt}
\renewcommand{\arraystretch}{1.08}

\newcommand{\surveyentry}[2]{#1~\citep{#2}}
\newcommand{\surveyfocus}[1]{\mbox{#1}}

\renewcommand{\tabularxcolumn}[1]{m{#1}}
\newcolumntype{Y}{>{\centering\arraybackslash}X}
\newcolumntype{Q}[1]{>{\raggedright\arraybackslash}m{#1}}

\caption{\textbf{Positioning related surveys.}
Checks mark dimensions central to a survey; crosses mark those outside it.
Six of the ten surveys cover both computer-use and robot-use and four analyze benchmarks, yet none compares the two settings as one capability structure.
\papav{} adds that digital--physical analogy as an explicit axis.}
\label{tab:survey-position}

\rowcolors{2}{slTint!45}{subcategory}
\begin{tabularx}{\linewidth}{
    l
    !{\color{tabledivider}\vrule width 0.5pt}
    Q{0.32\linewidth}
    !{\color{tabledivider}\vrule width 0.5pt}
    *{4}{Y}
}
\toprule
\rowcolor{slTint}
\textbf{Survey}
& \makebox[\linewidth][c]{%
    \makecell[c]{\textbf{Primary}\\\textbf{Focus}}}
& \makecell[c]{\textbf{Computer-}\\\textbf{Use}}
& \makecell[c]{\textbf{Robot-}\\\textbf{Use}}
& \makecell[c]{\textbf{Benchmark}\\\textbf{Analysis}}
& \makecell[c]{\textbf{Digit.--Phys.}\\\textbf{Analogy}}
\\
\midrule

\surveyentry{Agent AI}{durante2024agentaisurveyinghorizons}
& \surveyfocus{Multimodal interaction}
& \surveyfull & \surveyfull & \surveynone & \surveynone
\\

\surveyentry{Multimodal Agents}{xie2024largemultimodalagentssurvey}
& \surveyfocus{Multimodal agent architecture}
& \surveyfull & \surveynone & \surveyfull & \surveynone
\\

\surveyentry{Embodied AI}{liu2025aligningcyberspacephysical}
& \surveyfocus{Multimodal models for embodiment}
& \surveyfull & \surveyfull & \surveynone & \surveynone
\\

\surveyentry{GUI Agents}{tang2025surveymllmbasedguiagents}
& \surveyfocus{(M)LLM-based GUI Agents}
& \surveyfull & \surveynone & \surveyfull & \surveynone
\\

\surveyentry{VLA Models}{ma2026surveyvisionlanguageactionmodelsembodied}
& \surveyfocus{VLA-based embodied AI}
& \surveynone & \surveyfull & \surveyfull & \surveynone
\\

\surveyentry{Robot Agents}{salimpour2025embodiedagenticaireview}
& \surveyfocus{LLM/VLM-driven robot autonomy}
& \surveyfull & \surveyfull & \surveynone & \surveynone
\\

\surveyentry{Agentic MLLMs}{yao2025surveyagenticmultimodallarge}
& \surveyfocus{Agentic multimodal LLMs}
& \surveyfull & \surveyfull & \surveynone & \surveynone
\\

\surveyentry{Agentic AI}{v2026agenticartificialintelligenceai}
& \surveyfocus{Autonomous LLM agents}
& \surveyfull & \surveyfull & \surveynone & \surveynone
\\

\surveyentry{Agent Frameworks}{mokaria2026survey}
& \surveyfocus{Multimodal agentic frameworks}
& \surveyfull & \surveyfull & \surveyfull & \surveynone
\\

\surveyentry{LMM-Based Robots}{tu2026large}
& \surveyfocus{LMM-based embodied robots}
& \surveynone & \surveyfull & \surveynone & \surveynone
\\

\midrule
\rowcolor{lightgreen}
\textbf{\papav{} (Ours)}
& \surveyfocus{Capabilities from computer-use to robot-use}
& \surveyfull & \surveyfull & \surveyfull & \surveyfull
\\

\bottomrule
\end{tabularx}
\end{table}

%% file: figures/taxonomy/taxonomy.tex
% Generated by draw_graph.py; edit capabilities.yaml/style.json, not this file.
% Main-document preamble requirements:
%   \usepackage{graphicx}
%   \usepackage{forest}
%   \usepackage{xcolor}
%   \usepackage{natbib}
%   \usepackage[colorlinks=true]{hyperref}
% Include this figure with \input{taxonomy.tex}, not \includegraphics.
\providecommand{\taxonomysectionlink}[2]{%
  \ifcsname r@#1\endcsname%
    \hyperref[#1]{(\S{}\ref*{#1})}%
  \else%
    (\S{}#2)%
  \fi%
}%
% Rotate vertical parent titles and numbers separately.  The hyperlink
% encloses only the already-rotated number, so its PDF click area remains
% valid while the title itself stays non-clickable.
\providecommand{\taxonomyverticalsectionlabel}[3]{%
  \shortstack[c]{%
    \ifcsname r@#1\endcsname%
      \hyperref[#1]{\rotatebox{90}{(\S{}\ref*{#1})}}%
    \else%
      \rotatebox{90}{(\S{}#2)}%
    \fi%
    \\[-2pt]%
    \rotatebox{90}{#3}%
  }%
}%
% Make the displayed work name itself link to the bibliography.
% hidelinks preserves the surrounding MMEA/MMA/RS text color, while
% nocite registers the BibTeX key without printing a visible [x].
\providecommand{\taxonomyworklink}[2]{%
  \nocite{#1}%
  \begingroup%
  \hypersetup{hidelinks}%
  \hyperlink{cite.#1}{#2}%
  \endgroup%
}%
\definecolor{taxonomyPaperClassMmeaText}{HTML}{439753}%
\definecolor{taxonomyPaperClassMmaText}{HTML}{3377D1}%
\definecolor{taxonomyPaperClassRsText}{HTML}{747A3F}%
\definecolor{taxonomy0Fill}{HTML}{FFFFFF}%
\definecolor{taxonomy0Draw}{HTML}{444444}%
\definecolor{taxonomy1Fill}{HTML}{DDEBF7}%
\definecolor{taxonomy1Draw}{HTML}{5B9BD5}%
\definecolor{taxonomy2Fill}{HTML}{EAF2F8}%
\definecolor{taxonomy2Draw}{HTML}{7FAED3}%
\definecolor{taxonomy3Fill}{HTML}{F4F8FB}%
\definecolor{taxonomy3Draw}{HTML}{9ABDD8}%
\definecolor{taxonomyPapersFill}{HTML}{FFFFFF}%
\definecolor{taxonomyPapersDraw}{HTML}{BBBBBB}%
\definecolor{taxonomynode_1Fill}{HTML}{FFFFFF}%
\definecolor{taxonomynode_1Draw}{HTML}{444444}%
\definecolor{taxonomynode_2Fill}{HTML}{F4F8FB}%
\definecolor{taxonomynode_2Draw}{HTML}{5B9BD5}%
\definecolor{taxonomynode_3Fill}{HTML}{FFF5F4}%
\definecolor{taxonomynode_3Draw}{HTML}{F57C6E}%
\definecolor{taxonomynode_3Text}{HTML}{B84438}%
\definecolor{taxonomynode_4Fill}{HTML}{FFFFFF}%
\definecolor{taxonomynode_4Draw}{HTML}{B7CCE2}%
\definecolor{taxonomynode_5Fill}{HTML}{FEF9F4}%
\definecolor{taxonomynode_5Draw}{HTML}{F2B56F}%
\definecolor{taxonomynode_5Text}{HTML}{A96924}%
\definecolor{taxonomynode_6Fill}{HTML}{FFFFFF}%
\definecolor{taxonomynode_6Draw}{HTML}{B7CCE2}%
\definecolor{taxonomynode_7Fill}{HTML}{F6FAFA}%
\definecolor{taxonomynode_7Draw}{HTML}{84C3B7}%
\definecolor{taxonomynode_7Text}{HTML}{3F776E}%
\definecolor{taxonomynode_8Fill}{HTML}{FFFFFF}%
\definecolor{taxonomynode_8Draw}{HTML}{B7CCE2}%
\definecolor{taxonomynode_9Fill}{HTML}{F5FAFE}%
\definecolor{taxonomynode_9Draw}{HTML}{71B7ED}%
\definecolor{taxonomynode_9Text}{HTML}{2F73A6}%
\definecolor{taxonomynode_10Fill}{HTML}{FFFFFF}%
\definecolor{taxonomynode_10Draw}{HTML}{B7CCE2}%
\definecolor{taxonomynode_11Fill}{HTML}{FAF9FE}%
\definecolor{taxonomynode_11Draw}{HTML}{B8AEEB}%
\definecolor{taxonomynode_11Text}{HTML}{655A9B}%
\definecolor{taxonomynode_12Fill}{HTML}{FFFFFF}%
\definecolor{taxonomynode_12Draw}{HTML}{B7CCE2}%
\definecolor{taxonomynode_13Fill}{HTML}{F4F8FB}%
\definecolor{taxonomynode_13Draw}{HTML}{8064A2}%
\definecolor{taxonomynode_14Fill}{HTML}{FCFEFD}%
\definecolor{taxonomynode_14Draw}{HTML}{CFEEDC}%
\definecolor{taxonomynode_14Text}{HTML}{439753}%
\definecolor{taxonomynode_15Fill}{HTML}{FFFFFF}%
\definecolor{taxonomynode_15Draw}{HTML}{C8BDD8}%
\definecolor{taxonomynode_16Fill}{HTML}{FAFFFF}%
\definecolor{taxonomynode_16Draw}{HTML}{B7F0F0}%
\definecolor{taxonomynode_16Text}{HTML}{3377D1}%
\definecolor{taxonomynode_17Fill}{HTML}{FFFFFF}%
\definecolor{taxonomynode_17Draw}{HTML}{C8BDD8}%
\definecolor{taxonomynode_18Fill}{HTML}{FDFEFB}%
\definecolor{taxonomynode_18Draw}{HTML}{E7ECC8}%
\definecolor{taxonomynode_18Text}{HTML}{747A3F}%
\definecolor{taxonomynode_19Fill}{HTML}{FFFFFF}%
\definecolor{taxonomynode_19Draw}{HTML}{C8BDD8}%
% Measure only Capabilities rows with the font of the including document.
\newlength{\taxonomyPaperWidth}%
\newlength{\taxonomyPaperCandidateWidth}%
\newcommand{\taxonomymeasurepaperrow}[1]{%
  \settowidth{\taxonomyPaperCandidateWidth}{{\normalfont\fontsize{9}{10.8}\selectfont #1}}%
  \ifdim\taxonomyPaperCandidateWidth>\taxonomyPaperWidth%
    \setlength{\taxonomyPaperWidth}{\taxonomyPaperCandidateWidth}%
  \fi%
}%
\taxonomymeasurepaperrow{\textbf{MMEA [}Explore-EQA, RB-VLA, 3D-Belief, Gemini Robotics, EMOS\textbf{]}}%
\taxonomymeasurepaperrow{\textbf{MMA [}MM-REACT, M3-Agent, SeeClick, Avis, VideoAgent, OmniAgent\textbf{]}}%
\taxonomymeasurepaperrow{\textbf{RS [}SeeHearFeel, Kimera, APPLE, TOAPP-EPKS, OK-Robot, HOV-SG\textbf{]}}%
\taxonomymeasurepaperrow{\textbf{MMEA [}3D-Belief, SayCan, Reflective Planning, VoxPoser, WLA-0\textbf{]}}%
\taxonomymeasurepaperrow{\textbf{MMA [}AppDeltaWorld, WMA, MobileDreamer, WebDreamer, Dynalang\textbf{]}}%
\taxonomymeasurepaperrow{\textbf{RS [}VPP, DayDreamer, ModularAgent, PlaNet, DeepVis, DreamerV1\textbf{]}}%
\taxonomymeasurepaperrow{\textbf{MMEA [}ProgPrompt, Inner Monologue, KnowNo, RoCo, RoboAgent\textbf{]}}%
\taxonomymeasurepaperrow{\textbf{MMA [}WebOperator, ControlLLM, BacktrackAgent, HuggingGPT, Llava\textbf{]}}%
\taxonomymeasurepaperrow{\textbf{RS [}Pddlstream, Hierarchical, CaStL, SayPlan, RoboDreamer, HPT\textbf{]}}%
\taxonomymeasurepaperrow{\textbf{MMEA [}Gemini Robotics, Hi Robot, VoxPoser,Code as Policies, RB-VLA\textbf{]}}%
\taxonomymeasurepaperrow{\textbf{MMA [}GPT-4V, Copilot, ShowUI, STEVE-1, SeeClick, Aguvis, CogAgent\textbf{]}}%
\taxonomymeasurepaperrow{\textbf{RS [}RT-2, Act, Diffusion Policy, RDT-1B, Pi0, FAST, VIMA, RT-X, A2C2\textbf{]}}%
\taxonomymeasurepaperrow{\textbf{MMEA [}Reflect, DoReMi, Inner Monologue, Code-as-Monitor, PaLM-E\textbf{]}}%
\taxonomymeasurepaperrow{\textbf{MMA [}Voyager, MobileUse, VAGEN, IRA, GUISpector, BacktrackAgent\textbf{]}}%
\taxonomymeasurepaperrow{\textbf{RS [}Rl-vlm-f, Liv, SAFE, FAIL-Detect, CheckVLA, RoboOS, ActProbe\textbf{]}}%
\definecolor{taxonomyEdge}{HTML}{777777}%
\begin{forest}
  taxonomy base/.style={
    rectangle, rounded corners=5pt, align=left,
    line width=1.20pt,
    inner xsep=8.64pt, inner ysep=3.24pt,
    font=\normalfont\fontsize{10}{12}\selectfont
  },
  taxonomy level 0/.style={taxonomy base,
    fill=taxonomy0Fill,
    draw=taxonomy0Draw,
  },
  taxonomy level 1/.style={taxonomy base,
    fill=taxonomy1Fill,
    draw=taxonomy1Draw,
  },
  taxonomy level 2/.style={taxonomy base,
    fill=taxonomy2Fill,
    draw=taxonomy2Draw,
  },
  taxonomy level 3/.style={taxonomy base,
    fill=taxonomy3Fill,
    draw=taxonomy3Draw,
  },
  taxonomy level papers/.style={taxonomy base,
    fill=taxonomyPapersFill,
    draw=taxonomyPapersDraw,
  },
  taxonomy level papers/.append style={
    line width=0.80pt,
    inner xsep=3.64pt,
    font=\normalfont\fontsize{9}{10.8}\selectfont
  },
  for tree={
    grow'=east, anchor=west, parent anchor=east, child anchor=west,
    l sep=18pt, s sep=8pt,
    edge={draw=taxonomyEdge, line width=1.00pt},
    edge path={
      \noexpand\path[\forestoption{edge}]
      (!u.parent anchor) -- ++(7pt,0) |- (.child anchor)
      \forestoption{edge label};
    },
  },
  [{\rotatebox{90}{Research Landscape}}, taxonomy level 0, fill=taxonomynode_1Fill, draw=taxonomynode_1Draw, font=\normalfont\bfseries\fontsize{10}{12}\selectfont
    [{\taxonomyverticalsectionlabel{sec:capabilities}{3}{Capabilities}}, taxonomy level 1, fill=taxonomynode_2Fill, draw=taxonomynode_2Draw, font=\normalfont\bfseries\fontsize{10}{12}\selectfont
      [{Perceive~\taxonomysectionlink{sec:perceive}{3.1}}, taxonomy level 2, fill=taxonomynode_3Fill, draw=taxonomynode_3Draw, font=\normalfont\bfseries\fontsize{10}{12}\selectfont, l sep=18pt, minimum width=96pt, align=center, text=taxonomynode_3Text
        [{\parbox[t]{\taxonomyPaperWidth}{\raggedright \mbox{\textcolor{taxonomyPaperClassMmeaText}{\textbf{MMEA [}}\mbox{\textcolor{black}{\taxonomyworklink{exploreconfident2024}{Explore-EQA}}}, \mbox{\textcolor{black}{\taxonomyworklink{rbvla2026}{RB-VLA}}}, \mbox{\textcolor{black}{\taxonomyworklink{threedbelief2026}{3D-Belief}}}, \mbox{\textcolor{black}{\taxonomyworklink{gemini2025robotics}{Gemini Robotics}}}, \mbox{\textcolor{black}{\taxonomyworklink{chen2025emos}{EMOS}}}\textcolor{taxonomyPaperClassMmeaText}{\textbf{]}}}\\ \mbox{\textcolor{taxonomyPaperClassMmaText}{\textbf{MMA [}}\mbox{\textcolor{black}{\taxonomyworklink{yang2023mm}{MM-REACT}}}, \mbox{\textcolor{black}{\taxonomyworklink{long2026seeing}{M3-Agent}}}, \mbox{\textcolor{black}{\taxonomyworklink{cheng2024seeclick}{SeeClick}}}, \mbox{\textcolor{black}{\taxonomyworklink{hu2023avis}{Avis}}}, \mbox{\textcolor{black}{\taxonomyworklink{fan2024videoagent}{VideoAgent}}}, \mbox{\textcolor{black}{\taxonomyworklink{tao2025active}{OmniAgent}}}\textcolor{taxonomyPaperClassMmaText}{\textbf{]}}}\\ \mbox{\textcolor{taxonomyPaperClassRsText}{\textbf{RS [}}\mbox{\textcolor{black}{\taxonomyworklink{li2022see}{SeeHearFeel}}}, \mbox{\textcolor{black}{\taxonomyworklink{rosinol2020kimera}{Kimera}}}, \mbox{\textcolor{black}{\taxonomyworklink{schneider2026apple}{APPLE}}}, \mbox{\textcolor{black}{\taxonomyworklink{ghasemi2020task}{TOAPP-EPKS}}}, \mbox{\textcolor{black}{\taxonomyworklink{liu2024okrobot}{OK-Robot}}}, \mbox{\textcolor{black}{\taxonomyworklink{werby2024hierarchical}{HOV-SG}}}\textcolor{taxonomyPaperClassRsText}{\textbf{]}}}}}, taxonomy level papers, fill=taxonomynode_4Fill, draw=taxonomynode_4Draw
        ]
      ]
      [{Anticipate~\taxonomysectionlink{sec:anticipate}{3.2}}, taxonomy level 2, fill=taxonomynode_5Fill, draw=taxonomynode_5Draw, font=\normalfont\bfseries\fontsize{10}{12}\selectfont, l sep=18pt, minimum width=96pt, align=center, text=taxonomynode_5Text
        [{\parbox[t]{\taxonomyPaperWidth}{\raggedright \mbox{\textcolor{taxonomyPaperClassMmeaText}{\textbf{MMEA [}}\mbox{\textcolor{black}{\taxonomyworklink{threedbelief2026}{3D-Belief}}}, \mbox{\textcolor{black}{\taxonomyworklink{ahn2022can}{SayCan}}}, \mbox{\textcolor{black}{\taxonomyworklink{reflectiveplanning2025}{Reflective Planning}}}, \mbox{\textcolor{black}{\taxonomyworklink{huang2023voxposer}{VoxPoser}}}, \mbox{\textcolor{black}{\taxonomyworklink{wla2026}{WLA-0}}}\textcolor{taxonomyPaperClassMmeaText}{\textbf{]}}}\\ \mbox{\textcolor{taxonomyPaperClassMmaText}{\textbf{MMA [}}\mbox{\textcolor{black}{\taxonomyworklink{xu2026appdeltaworld}{AppDeltaWorld}}}, \mbox{\textcolor{black}{\taxonomyworklink{chae2025web}{WMA}}}, \mbox{\textcolor{black}{\taxonomyworklink{cao2026mobiledreamer}{MobileDreamer}}}, \mbox{\textcolor{black}{\taxonomyworklink{gu2024your}{WebDreamer}}}, \mbox{\textcolor{black}{\taxonomyworklink{lin2023learning}{Dynalang}}}\textcolor{taxonomyPaperClassMmaText}{\textbf{]}}}\\ \mbox{\textcolor{taxonomyPaperClassRsText}{\textbf{RS [}}\mbox{\textcolor{black}{\taxonomyworklink{hu2024video}{VPP}}}, \mbox{\textcolor{black}{\taxonomyworklink{pmlr-v205-wu23c}{DayDreamer}}}, \mbox{\textcolor{black}{\taxonomyworklink{zhan2026modularagent}{ModularAgent}}}, \mbox{\textcolor{black}{\taxonomyworklink{hafner2019planet}{PlaNet}}}, \mbox{\textcolor{black}{\taxonomyworklink{finn2017deep}{DeepVis}}}, \mbox{\textcolor{black}{\taxonomyworklink{hafner2019dream}{DreamerV1}}}\textcolor{taxonomyPaperClassRsText}{\textbf{]}}}}}, taxonomy level papers, fill=taxonomynode_6Fill, draw=taxonomynode_6Draw
        ]
      ]
      [{Plan~\taxonomysectionlink{sec:plan}{3.3}}, taxonomy level 2, fill=taxonomynode_7Fill, draw=taxonomynode_7Draw, font=\normalfont\bfseries\fontsize{10}{12}\selectfont, l sep=18pt, minimum width=96pt, align=center, text=taxonomynode_7Text
        [{\parbox[t]{\taxonomyPaperWidth}{\raggedright \mbox{\textcolor{taxonomyPaperClassMmeaText}{\textbf{MMEA [}}\mbox{\textcolor{black}{\taxonomyworklink{singh2023progprompt}{ProgPrompt}}}, \mbox{\textcolor{black}{\taxonomyworklink{huang2022inner}{Inner Monologue}}}, \mbox{\textcolor{black}{\taxonomyworklink{ren2023knowno}{KnowNo}}}, \mbox{\textcolor{black}{\taxonomyworklink{mandi2024roco}{RoCo}}}, \mbox{\textcolor{black}{\taxonomyworklink{xu2026roboagent}{RoboAgent}}}\textcolor{taxonomyPaperClassMmeaText}{\textbf{]}}}\\ \mbox{\textcolor{taxonomyPaperClassMmaText}{\textbf{MMA [}}\mbox{\textcolor{black}{\taxonomyworklink{weboperator2025}{WebOperator}}}, \mbox{\textcolor{black}{\taxonomyworklink{liu2024controlllm}{ControlLLM}}}, \mbox{\textcolor{black}{\taxonomyworklink{wu2025backtrackagent}{BacktrackAgent}}}, \mbox{\textcolor{black}{\taxonomyworklink{shen2023hugginggpt}{HuggingGPT}}}, \mbox{\textcolor{black}{\taxonomyworklink{liu2024llava}{Llava}}}\textcolor{taxonomyPaperClassMmaText}{\textbf{]}}}\\ \mbox{\textcolor{taxonomyPaperClassRsText}{\textbf{RS [}}\mbox{\textcolor{black}{\taxonomyworklink{garrett2020pddlstream}{Pddlstream}}}, \mbox{\textcolor{black}{\taxonomyworklink{kaelbling2011hierarchical}{Hierarchical}}}, \mbox{\textcolor{black}{\taxonomyworklink{castl2024}{CaStL}}}, \mbox{\textcolor{black}{\taxonomyworklink{sayplan2023}{SayPlan}}}, \mbox{\textcolor{black}{\taxonomyworklink{robodreamer2024}{RoboDreamer}}}, \mbox{\textcolor{black}{\taxonomyworklink{wang2024hpt}{HPT}}}\textcolor{taxonomyPaperClassRsText}{\textbf{]}}}}}, taxonomy level papers, fill=taxonomynode_8Fill, draw=taxonomynode_8Draw
        ]
      ]
      [{Act~\taxonomysectionlink{sec:act}{3.4}}, taxonomy level 2, fill=taxonomynode_9Fill, draw=taxonomynode_9Draw, font=\normalfont\bfseries\fontsize{10}{12}\selectfont, l sep=18pt, minimum width=96pt, align=center, text=taxonomynode_9Text
        [{\parbox[t]{\taxonomyPaperWidth}{\raggedright \mbox{\textcolor{taxonomyPaperClassMmeaText}{\textbf{MMEA [}}\mbox{\textcolor{black}{\taxonomyworklink{gemini2025robotics}{Gemini Robotics}}}, \mbox{\textcolor{black}{\taxonomyworklink{shi2025hi}{Hi Robot}}}, \mbox{\textcolor{black}{\taxonomyworklink{huang2023voxposer}{VoxPoser}}}, \mbox{\textcolor{black}{\taxonomyworklink{liang2023code}{Code as Policies}}}, \mbox{\textcolor{black}{\taxonomyworklink{rbvla2026}{RB-VLA}}}\textcolor{taxonomyPaperClassMmeaText}{\textbf{]}}}\\ \mbox{\textcolor{taxonomyPaperClassMmaText}{\textbf{MMA [}}\mbox{\textcolor{black}{\taxonomyworklink{zheng2024gpt}{GPT-4V}}}, \mbox{\textcolor{black}{\taxonomyworklink{wu2024copilot}{Copilot}}}, \mbox{\textcolor{black}{\taxonomyworklink{shou2025showui}{ShowUI}}}, \mbox{\textcolor{black}{\taxonomyworklink{lifshitz2023steve}{STEVE-1}}}, \mbox{\textcolor{black}{\taxonomyworklink{cheng2024seeclick}{SeeClick}}}, \mbox{\textcolor{black}{\taxonomyworklink{xu2024aguvis}{Aguvis}}}, \mbox{\textcolor{black}{\taxonomyworklink{hong2024cogagent}{CogAgent}}}\textcolor{taxonomyPaperClassMmaText}{\textbf{]}}}\\ \mbox{\textcolor{taxonomyPaperClassRsText}{\textbf{RS [}}\mbox{\textcolor{black}{\taxonomyworklink{brohan2023rt2}{RT-2}}}, \mbox{\textcolor{black}{\taxonomyworklink{zhao2023act}{Act}}}, \mbox{\textcolor{black}{\taxonomyworklink{chi2023diffusionpolicy}{Diffusion Policy}}}, \mbox{\textcolor{black}{\taxonomyworklink{liu2024rdt}{RDT-1B}}}, \mbox{\textcolor{black}{\taxonomyworklink{black2024pi0}{Pi0}}}, \mbox{\textcolor{black}{\taxonomyworklink{fast2025}{FAST}}}, \mbox{\textcolor{black}{\taxonomyworklink{jiang2023vima}{VIMA}}}, \mbox{\textcolor{black}{\taxonomyworklink{oxe2024}{RT-X}}}, \mbox{\textcolor{black}{\taxonomyworklink{a2c2_2025}{A2C2}}}\textcolor{taxonomyPaperClassRsText}{\textbf{]}}}}}, taxonomy level papers, fill=taxonomynode_10Fill, draw=taxonomynode_10Draw
        ]
      ]
      [{Verify~\taxonomysectionlink{sec:verify}{3.5}}, taxonomy level 2, fill=taxonomynode_11Fill, draw=taxonomynode_11Draw, font=\normalfont\bfseries\fontsize{10}{12}\selectfont, l sep=18pt, minimum width=96pt, align=center, text=taxonomynode_11Text
        [{\parbox[t]{\taxonomyPaperWidth}{\raggedright \mbox{\textcolor{taxonomyPaperClassMmeaText}{\textbf{MMEA [}}\mbox{\textcolor{black}{\taxonomyworklink{liu2023reflect}{Reflect}}}, \mbox{\textcolor{black}{\taxonomyworklink{guo2024doremi}{DoReMi}}}, \mbox{\textcolor{black}{\taxonomyworklink{huang2022inner}{Inner Monologue}}}, \mbox{\textcolor{black}{\taxonomyworklink{codeasmonitor2024}{Code-as-Monitor}}}, \mbox{\textcolor{black}{\taxonomyworklink{driess2023palm}{PaLM-E}}}\textcolor{taxonomyPaperClassMmeaText}{\textbf{]}}}\\ \mbox{\textcolor{taxonomyPaperClassMmaText}{\textbf{MMA [}}\mbox{\textcolor{black}{\taxonomyworklink{wang2024voyager}{Voyager}}}, \mbox{\textcolor{black}{\taxonomyworklink{li2026mobileuse}{MobileUse}}}, \mbox{\textcolor{black}{\taxonomyworklink{vagen2026}{VAGEN}}}, \mbox{\textcolor{black}{\taxonomyworklink{shi2026interactive}{IRA}}}, \mbox{\textcolor{black}{\taxonomyworklink{kolthoff2025guispector}{GUISpector}}}, \mbox{\textcolor{black}{\taxonomyworklink{wu2025backtrackagent}{BacktrackAgent}}}\textcolor{taxonomyPaperClassMmaText}{\textbf{]}}}\\ \mbox{\textcolor{taxonomyPaperClassRsText}{\textbf{RS [}}\mbox{\textcolor{black}{\taxonomyworklink{wang2024rl}{Rl-vlm-f}}}, \mbox{\textcolor{black}{\taxonomyworklink{ma2023liv}{Liv}}}, \mbox{\textcolor{black}{\taxonomyworklink{safe2025}{SAFE}}}, \mbox{\textcolor{black}{\taxonomyworklink{xu2025can}{FAIL-Detect}}}, \mbox{\textcolor{black}{\taxonomyworklink{checkvla2026}{CheckVLA}}}, \mbox{\textcolor{black}{\taxonomyworklink{tan2025roboos}{RoboOS}}}, \mbox{\textcolor{black}{\taxonomyworklink{actprobe2026}{ActProbe}}}\textcolor{taxonomyPaperClassRsText}{\textbf{]}}}}}, taxonomy level papers, fill=taxonomynode_12Fill, draw=taxonomynode_12Draw
        ]
      ]
    ]
    [{\taxonomyverticalsectionlabel{sec:benchmark}{4}{Benchmark}}, taxonomy level 1, fill=taxonomynode_13Fill, draw=taxonomynode_13Draw, calign=child, calign child=2, font=\normalfont\bfseries\fontsize{10}{12}\selectfont
      [{MMEA~\taxonomysectionlink{sec:MMEA}{4.1}}, taxonomy level 2, fill=taxonomynode_14Fill, draw=taxonomynode_14Draw, font=\normalfont\bfseries\fontsize{10}{12}\selectfont, l sep=18pt, minimum width=96pt, align=center, text=taxonomynode_14Text
        [{\parbox[t]{\taxonomyPaperWidth}{\raggedright \mbox{\textcolor{black}{\taxonomyworklink{hong2026esi}{ESI-Bench}}}, \mbox{\textcolor{black}{\taxonomyworklink{padmakumar2022teach}{TEACh}}}, \mbox{\textcolor{black}{\taxonomyworklink{mandi2024roco}{RoCoBench}}}, \mbox{\textcolor{black}{\taxonomyworklink{lu2026bench}{IS-Bench}}}, \mbox{\textcolor{black}{\taxonomyworklink{chen2025emos}{Habitat-MAS}}}, \mbox{\textcolor{black}{\taxonomyworklink{han2026robocerebra}{RoboCerebra}}}}}, taxonomy level papers, fill=taxonomynode_15Fill, draw=taxonomynode_15Draw
        ]
      ]
      [{MMA~\taxonomysectionlink{sec:MMA}{4.2}}, taxonomy level 2, fill=taxonomynode_16Fill, draw=taxonomynode_16Draw, font=\normalfont\bfseries\fontsize{10}{12}\selectfont, l sep=18pt, minimum width=96pt, align=center, text=taxonomynode_16Text
        [{\parbox[t]{\taxonomyPaperWidth}{\raggedright \mbox{\textcolor{black}{\taxonomyworklink{yang2025thinkinginspace}{VSI-Bench}}}, \mbox{\textcolor{black}{\taxonomyworklink{rawles2025androidworld}{AndroidWorld}}}, \mbox{\textcolor{black}{\taxonomyworklink{yue2025mmmu}{MMMU-Pro}}}, \mbox{\textcolor{black}{\taxonomyworklink{zheng2025agentstudio}{AgentStudio}}}, \mbox{\textcolor{black}{\taxonomyworklink{fu2026video}{Video-MME-v2}}}, \mbox{\textcolor{black}{\taxonomyworklink{tian2025mmina}{MMInA}}}, \mbox{\textcolor{black}{\taxonomyworklink{koh2024visualwebarena}{VisualWebArena}}}, \mbox{\textcolor{black}{\taxonomyworklink{xu2025crab}{CRAB}}}, \mbox{\textcolor{black}{\taxonomyworklink{zhou2024webarena}{WebArena}}}, \mbox{\textcolor{black}{\taxonomyworklink{yuan2026osworld2}{OSWorld2.0}}}, \mbox{\textcolor{black}{\taxonomyworklink{ouyang2026gameworld}{GameWorld}}}}}, taxonomy level papers, fill=taxonomynode_17Fill, draw=taxonomynode_17Draw
        ]
      ]
      [{RS~\taxonomysectionlink{sec:RS}{4.3}}, taxonomy level 2, fill=taxonomynode_18Fill, draw=taxonomynode_18Draw, font=\normalfont\bfseries\fontsize{10}{12}\selectfont, l sep=18pt, minimum width=96pt, align=center, text=taxonomynode_18Text
        [{\parbox[t]{\taxonomyPaperWidth}{\raggedright \mbox{\textcolor{black}{\taxonomyworklink{yenamandra2023homerobot}{HomeRobot}}}, \mbox{\textcolor{black}{\taxonomyworklink{jiang2026robowm}{RoboWM-Bench}}}, \mbox{\textcolor{black}{\taxonomyworklink{mu2025robotwin}{RoboTwin}}}, \mbox{\textcolor{black}{\taxonomyworklink{yang2026robolab}{RoboLab}}}, \mbox{\textcolor{black}{\taxonomyworklink{nasiriany2026robocasa365}{RoboCasa365}}}, \mbox{\textcolor{black}{\taxonomyworklink{li2023behavior}{BEHAVIOR-1K}}}, \mbox{\textcolor{black}{\taxonomyworklink{khanna2024goat}{GOAT-Bench}}}, \mbox{\textcolor{black}{\taxonomyworklink{chen2025robotwin}{RoboTwin 2.0}}}, \mbox{\textcolor{black}{\taxonomyworklink{shridhar2020alfred}{ALFRED}}}, \mbox{\textcolor{black}{\taxonomyworklink{liu2023libero}{LIBERO}}}, \mbox{\textcolor{black}{\taxonomyworklink{anderson2018vision}{R2R}}}}}, taxonomy level papers, fill=taxonomynode_19Fill, draw=taxonomynode_19Draw
        ]
      ]
    ]
  ]
\end{forest}

%% file: sections/2-background/background.tex
\section{Background and Problem Formulation}
\label{sec:background}

We adopt a task-level loop to compare multimodal agents (MMAs) and robotic systems (RSs). Each step contains one operation that advances the task, the evidence used to select it, and the feedback returned after it. In computer-use, this operation may be a GUI action or tool call. In robot-use, it may be a robot skill, waypoint, or trajectory command that a lower-level controller realizes over time. Working at the task level places both settings on a common time scale: a click and a control tick are not comparable, but a task step is. PAPAV describes five functional roles within this loop. It focuses on what each role contributes rather than how the system is implemented. This abstraction allows us to compare systems with different architectures and embodiments.

% ============================================================
\subsection{Multimodal Agents}
\label{sec:bg-mm-agent}

A multimodal agent is a goal-directed system that interacts with the environment through multimodal observations and external actions. Its observations may comprise language, images, structured states, and tool outputs. Its actions may involve language generation, GUI operations, API calls, or tool execution. At task step $k$, the agent selects an action according to $a_k \sim \pi(\cdot \mid g,h_k,m_k)$, where $g$ denotes the task goal, $h_k$ the interaction history, and $m_k$ the persistent memory. The action $a_k$ queries or modifies the external environment and yields a new observation $o_{k+1}$. The resulting observation is incorporated into the interaction history or persistent memory, providing context for subsequent decisions. This process forms a closed interaction loop in which the agent can adjust its behavior based on environmental feedback~\citep{yao2023react,yang2023mm,xie2024osworld,zheng2024synapse,shinn2023reflexion,wang2024voyager}.

% ============================================================
\subsection{Robotic Systems}
\label{sec:bg-robotic-system}

A robotic system interacts with the environment through sensing and actuation. Its observations may comprise vision, proprioception, force, and tactile signals. Its task-level actions may be represented as skills, waypoints, trajectories, or motion commands. These actions must be grounded with respect to the robot's embodiment and estimated state. Their execution is further subject to geometric, dynamic, safety, and control constraints~\citep{campos2021orb,garrett2021tamp,chen2026escaping,hogan1985impedance}.

At task step $k$, this grounding can be expressed as $a_k \xrightarrow{\mathrm{Exec}_e} u(t)$ for $t\in[t_k,t_{k+1})$. Here, $\mathrm{Exec}_e$ denotes the execution operator conditioned on embodiment $e$. The control input $u(t)$ may represent joint-position, velocity, or torque commands. This formulation also covers learned and end-to-end policies that implement the grounding process implicitly~\citep{brohan2023rt1,brohan2023rt2,black2024pi0, chen2026supervise}.

% ============================================================
\subsection{Multimodal Embodied Agents}
\label{sec:bg-mea}
A multimodal embodied agent (MMEA) can be characterized by five PAPAV capabilities: multimodal perception, consequence anticipation, task-level planning, embodied action, and outcome verification. The system maintains task state throughout sensing and execution. Execution feedback informs subsequent decisions. These functions operate as a persistent closed loop between multimodal task processing and physical execution. Observed effects may differ from anticipated outcomes during execution. These discrepancies can prompt the agent to revise its task belief, reassess action feasibility, and adjust the remaining plan. Failure of an individual action does not necessarily imply failure of the overall task. The agent can use available evidence to assess recovery options and select an alternative action consistent with the task goal. At step $k$, $h_k$ denotes the interaction history, $g$ the task goal, and $b_k$ the task-conditioned belief. The PAPAV loop can be formulated as:
\begin{equation}
(h_k,g) \xrightarrow{\mathsf{Perceive}} b_k \xrightarrow{\mathsf{Anticipate}} \widehat{\Delta}_k \xrightarrow{\mathsf{Plan}} p_k \xrightarrow{\mathsf{Act}} (a_k,\rho_k) \xrightarrow{T_d,O_d} o_{k+1} \xrightarrow{\mathsf{Verify}} v_k \circlearrowleft .
\label{eq:papav-loop}
\end{equation}
Here, $\widehat{\Delta}_k$ is the predicted task-relevant effect, $p_k$ the intended action course, $a_k$ the environment-facing action, and $\rho_k$ its execution progress. The operators $T_d$ and $O_d$ denote the transition and observation models of domain $d$. Verification produces a verdict $v_k$, such as continue, success, retry, replan, or fail. The observation and verdict provide input to the next interaction step.

We classify a system as an MMEA when multimodal task reasoning forms a persistent feedback loop around physical execution. The evaluated system configuration must satisfy three conditions:
\begin{enumerate}
    \item \textbf{Persistent embodied task loop:} task-relevant state persists across successive rounds of physical execution, and selected actions can produce observable changes in the physical environment.
    \item \textbf{Task-level capability returns:} each claimed PAPAV capability provides a documented output that can influence a subsequent task-level decision within the interaction loop.
    \item \textbf{Embodiment-grounded commitments:} task decisions explicitly account for physical feasibility, execution timing, and control constraints imposed by the deployed body and operating environment.
\end{enumerate}
A system is indeterminate when evidence for any condition is missing. An explicitly absent condition places the evaluated configuration outside the MMEA class. The criteria apply to complete system configurations and cover both modular and end-to-end implementations.

Embodiment limits the actions available at each task step. Given belief $b_k$ and embodiment $e$, the feasible action set can be expressed as:
\begin{equation}
\mathcal{A}_{\mathrm{feas}}(b_k,e)
=
\left\{
a\in\mathcal{A}
\,\middle|\,
\begin{aligned}
&\Gamma_j(a,b_k,e)\leq 0,
\quad j\in\{\mathrm{geo},\mathrm{dyn},\mathrm{safe}\},\\
&\tau_{\mathrm{exec}}(a,e)\leq \tau_{\max}
\end{aligned}
\right\}.
\label{eq:embodied-feasible-set}
\end{equation}
Here, $\mathcal{A}$ is the space of task-level interventions. The terms $\Gamma_{\mathrm{geo}}$, $\Gamma_{\mathrm{dyn}}$, and $\Gamma_{\mathrm{safe}}$ represent geometric feasibility, dynamic feasibility, and safety constraints, respectively. Geometric feasibility covers reachability and collision-free motion. Dynamic feasibility covers motion and actuator limits. Safety constraints capture task- and environment-specific risk bounds. An intervention satisfies a constraint when the corresponding value is non-positive. The term $\tau_{\mathrm{exec}}$ denotes the execution time required by the intervention, and $\tau_{\max}$ the time budget available at the current task step.

% ============================================================

\begin{figure}[!t]
  \centering
  \includegraphics[width=\linewidth,pagebox=cropbox,clip]{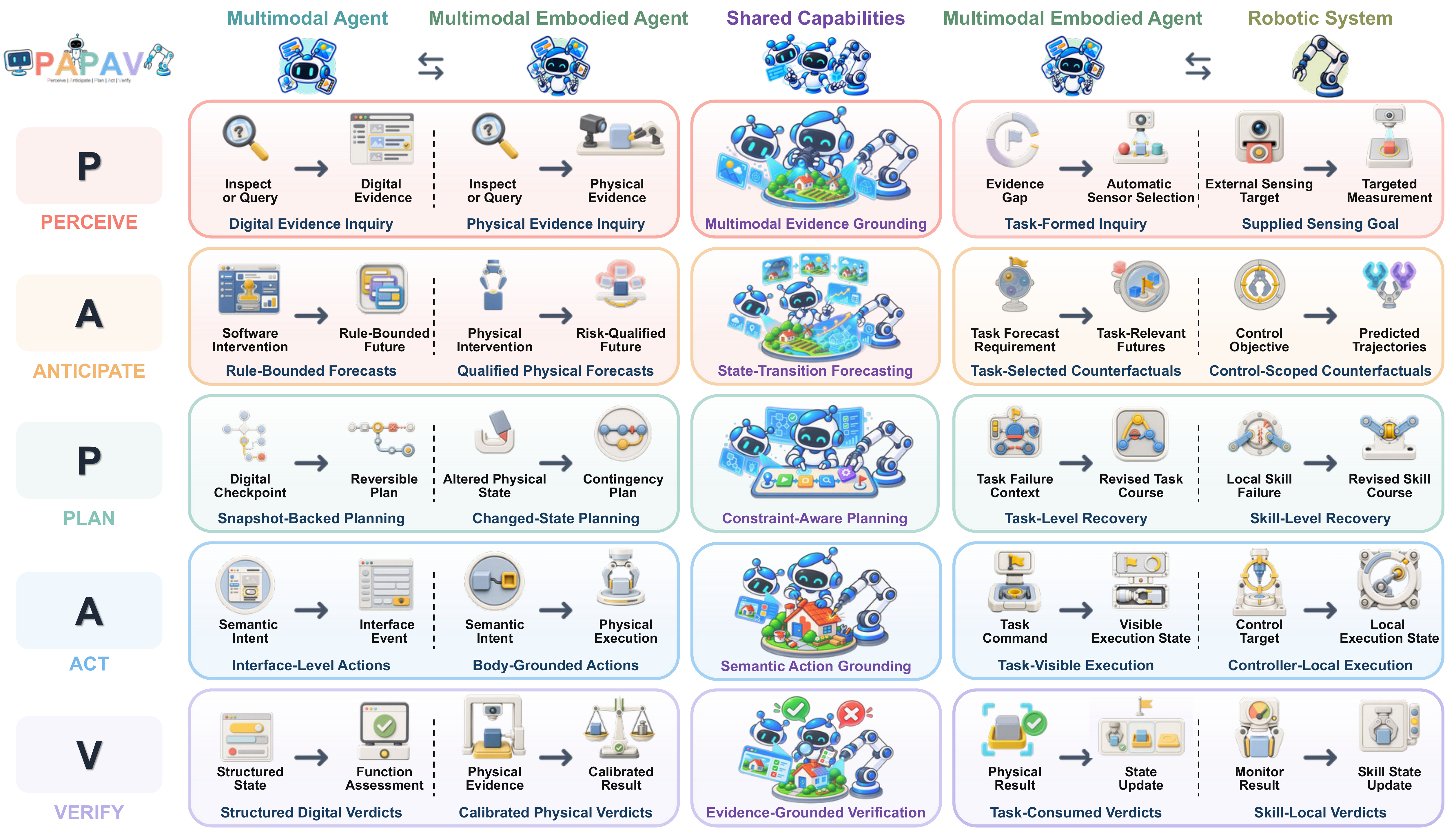}
  \caption{\textbf{PAPAV from computer-use to robot-use.} Each row contrasts how a shared capability is realized in multimodal agents, multimodal embodied agents, and robotic systems.}
  \label{fig:capabilities-main}
\end{figure}

\subsection{PAPAV Roles in the Task Loop}

Equation~\eqref{eq:papav-loop} defines five task-level roles by their outputs and uses. \textbf{Perceive} forms a task belief from multimodal evidence. \textbf{Anticipate} estimates the state changes caused by candidate interventions, and need only estimate decision-relevant consequences~\citep{pmlr-v205-wu23c}. \textbf{Plan} selects and organizes actions under task, environmental, and embodiment constraints. \textbf{Act} grounds the selected plan into an executable intervention and reports its progress. \textbf{Verify} converts post-action evidence into a verdict for continuation, termination, retry, or replanning, and must inform subsequent task decisions~\citep{huang2022inner,du2023successdetectors}. The roles may recur at different temporal resolutions, be combined or distributed across components, and jointly feed a single decision; throughout, we distinguish them by function and information flow.

%% file: sections/3-capability.tex
\section{Capabilities:
\textcolor{perceive}{P}%
\textcolor{anticipate}{A}%
\textcolor{plan}{P}%
\textcolor{act}{A}%
\textcolor{verify}{V}}
\label{sec:capabilities}

We use \papav{} %
as a unified capability framework to compare multimodal embodied agents (MMEAs), multimodal agents (MMAs), and robotic systems (RSs). Each capability is a recurring input--output relation within the task loop, defined by its role rather than by a module or a fixed execution order. This lets PAPAV trace the transition from computer-use to robot-use while preserving differences in implementation, embodiment, and environmental constraints. MMEAs lie at the intersection: physical embodiment constrains agent capabilities, while multimodal reasoning, goal-directed autonomy, and adaptive interaction extend the scope of conventional robotic systems.

Each subsection first identifies the capability shared by MMAs and RSs, then examines how MMEAs integrate and adapt it under physical constraints. Figure~\ref{fig:capabilities-main} serves as the roadmap: each row pairs a shared  capability with its MMEA--MMA and MMEA--RS comparisons. The MMEA, MMA, and RS labels in Figure~\ref{fig:taxonomy} mark non-exclusive evidence lineages.

\subsection{\textcolor{perceive}{P}erceive}
\label{sec:perceive}
\begin{figure}[!t]
  \centering
  \includegraphics[width=\linewidth,pagebox=cropbox,clip]{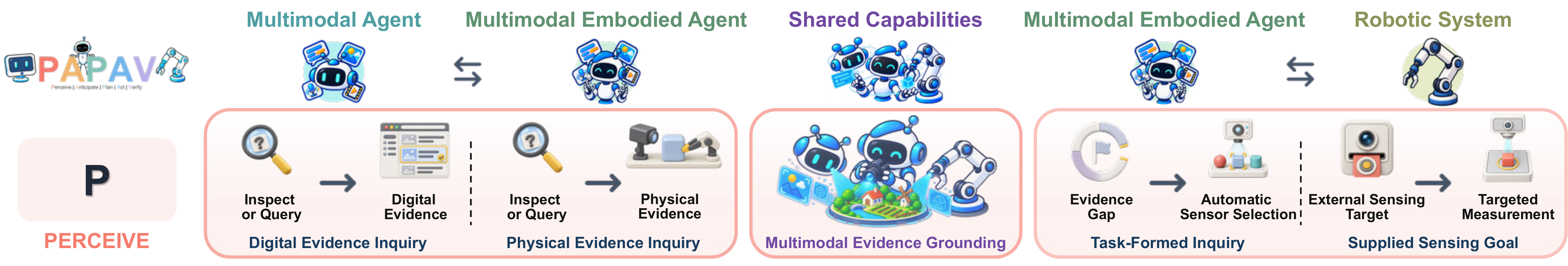}
  \caption{\textbf{Perceive: MMEA vs.\ MMA vs.\ RS.}
    Shared (center): multimodal evidence grounding. Left: digital evidence inquiry vs.\ embodied physical evidence acquisition. Right: sensing needs formed by the ongoing MMEA task vs.\ RS acquisition under supplied sensing objectives.}
  \label{fig:papav-perceive}
\end{figure}

Perception maps heterogeneous, partial observations and interaction history to an action-relevant belief about the environment, task, and agent. Across MMAs and RSs, state grounding and estimation exposes entities, relations, affordances, and uncertainties needed for the current objective and subsequent action. Under partial observability, an MMEA must maintain and revise this belief across interactions. Object-centric memories and agent-editable scene graphs retain state beyond the current view, while active-perception policies select observations~\citep{gu2024conceptgraphs,graphpad2025,activevla2026}. The agent preserves identity across viewpoints, answers open-vocabulary queries, records uncertainty and provenance, and revises outdated entries. Embodiment thus changes the responsibility, timing, and evidence for belief maintenance, while Verify separately tests whether goal-specific evidence supports a particular outcome judgment. Figure~\ref{fig:papav-perceive} compares evidence acquisition and the origin of sensing objectives.

\subsubsection{Perceive: Similarity}

\paragraph{From Incomplete Observations across Individual Modalities to Integrated Multimodal Evidence}
Each modality reveals different state variables and fails in different ways. Images contribute appearance and spatial cues; audio and video reveal events and temporal change; depth, proprioception, force, and touch expose geometry, embodiment, and contact. Multimodal agents fuse image regions, text, audio, video, interface state, and tool returns, while robots combine RGB, depth, proprioception, force, touch, and geometric cues. The task may also require temporal or spatial alignment, reconciliation of inconsistent signals, and retrieval of historical context~\citep{yang2023mm,shen2023hugginggpt,long2026seeing,lee2020multimodal,li2022see}. MM-REACT~\citep{yang2023mm} gathers image and video evidence from specialized tools; M3-Agent~\citep{long2026seeing} stores visual and auditory observations in entity-centered memory; See, Hear, and Feel~\citep{li2022see} joins global vision, acoustic events, and local tactile geometry for manipulation. The common relation is evidence integration: complementary sources recover task-relevant information that no single source supplies reliably.

\paragraph{From Observations to Grounded Environment-State Representations}
Integrated evidence remains insufficient for downstream reasoning unless it is grounded in an environment-state representation that specifies what currently exists and how it is changing. Perception must therefore associate observations with identifiable entities and events, estimate their current attributes, establish relations among them, and preserve their identities across time~\citep{ren2024timechat,peng2023openscene,werby2024hierarchical,gu2024conceptgraphs,long2026seeing}. The resulting representation may describe objects, speakers, video events, temporal intervals, digital entities, object poses, spatial configurations, motion, contact, or robot state. Although these variables differ across environments, their representations commonly encode identity, current state, relational structure, and temporal persistence~\citep{peng2023kosmos,tang2024salmonn,tao2025active,rosinol2020kimera,hughes2022hydra,gu2024conceptgraphs}. SeeClick~\citep{cheng2024seeclick} grounds language in screen coordinates; OSWorld~\citep{xie2024osworld} requires agents to recover application and interaction states from visual observations; and Kimera~\citep{rosinol2020kimera} converts continuous visual--inertial measurements into a dynamic scene graph containing persistent objects, agents, places, and spatiotemporal relations.

These mechanisms define Perceive independently of any sensor suite: available evidence is integrated and grounded in a task-relevant state representation~\citep{yang2023mm,li2022see,gu2024conceptgraphs}. What they leave unresolved is how to seek missing evidence when current observations cannot recover the required state. Embodiment turns observation into a situated, constrained interaction~\citep{schneider2026apple}. The comparison must therefore track who formulates the question, who controls acquisition, and how the returned evidence affects later decisions. The two contrasts below examine digital evidence requests~\citep{hu2023avis} and task-conditioned robotic active perception~\citep{exploreconfident2024,ghasemi2020task} from this task-level perspective.

\subsubsection{Perceive: Difference}

\paragraph{MMA vs. MMEA: From Digital Evidence Inquiry to Physical Evidence Inquiry}
Information seeking already forms part of the MMA decision loop. In AVIS~\citep{hu2023avis}, tool
selection follows from earlier results, returned evidence is interpreted by a
reasoner, and working memory carries that evidence into subsequent search. Embodiment relocates the inquiry to the agent's own
sensorimotor relation with the environment. A pending physical commitment may
expose an explicit question or an unresolved state variable, while
viewpoint, contact, timing, energy, and disturbance constrain how the uncertainty
can be resolved. PAPAV records the return as a
\emph{qualified belief}: a task-relevant proposition or state estimate accompanied
by the available confidence, provenance, acquisition conditions, and freshness.
The fields are analytical criteria, not properties already reported by every
cited system. An update becomes task-significant when it changes the plan,
the next capability invocation, or the conditions under which physical execution
may proceed within the current task loop.

\paragraph{RS vs. MMEA: From Supplied Sensing Goal to Task-Formed Inquiry}
Robotic active perception ranges from general acquisition policies to
task-conditioned perception; task relevance alone does not distinguish an
MMEA. APPLE~\citep{schneider2026apple} learns policies across several active-perception problems, while
task-oriented perception and planning~\citep{ghasemi2020task} update beliefs over task propositions and
adapt the associated plan. Explore until
Confident~\citep{exploreconfident2024} provides task-integrated embodied evidence: it receives an embodied
question, builds a semantic map, selects exploratory motion, and uses calibrated
confidence to decide when enough evidence has been collected.
The relevant distinction is functional. A persistent task process must issue or
revise the information need and later use the qualified return when choosing a
capability or physical commitment. A sensing specialist may still control camera
poses, tactile trajectories, stopping, and acquisition cost. The trace supports
the Perceive relation, whereas whole-system membership requires separate evidence.

\subsubsection{Perceive: Summary}

Perceive closes a task-level information gap with a belief revision that can
alter a later decision. Capability coding should trace four links: the
information need, embodied acquisition, qualified return, and consuming task
decision. These links are observability requirements for analysis, not a common
interface attributed to existing systems. Suitable traces can measure
calibration, freshness, acquisition cost, and evidence-induced decision changes.
Value of information is different because it asks what would have happened
without the acquired evidence; estimating it requires paired trials, branching
replay, or a validated simulator. An active-perception module may therefore
belong to either the robotics or embodied-agent evidence lineage. Whether it is
task-integrated depends on the route through which the resulting belief reaches
and changes the ongoing task process.

\subsection{\textcolor{anticipate}{A}nticipate}
\label{sec:anticipate}
\begin{figure}[!t]
  \centering
  \includegraphics[width=\linewidth,pagebox=cropbox,clip]{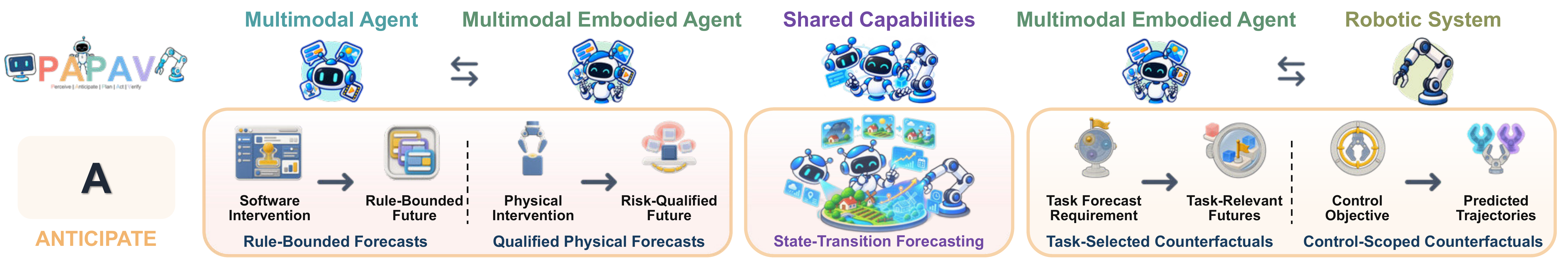}
  \caption{\textbf{Anticipate: MMEA vs.\ MMA vs.\ RS.}
    Shared (center): action-conditioned state-transition forecasting. Left: relatively rule-bounded digital futures vs.\ physical forecasts qualified by uncertainty, risk, and applicability. Right: physical counterfactuals selected for the current task vs.\ control-objective-scoped predictions.}
  \label{fig:papav-anticipate}
\end{figure}

Anticipation estimates consequences of a candidate action before commitment. Given belief $b_t$ and action $a_t$, action-conditioned prospective modeling predicts future states, observations, rewards, risks, or task progress. It answers the counterfactual ``what follows from this action?'', distinguishing action consequences from unconditional forecasts. Under embodiment, limited simulatability leaves an MMEA without a complete, authoritative simulator for open-ended physical interaction. Predictions span pixels, multi-view video, latent tokens, object state, contact, force, and task progress~\citep{irasim2024,genieenvisioner2025,interceptfuture2026,contactworld2026,tacforesight2026,kairos2026}. A useful forecast must distinguish candidate actions and inform planning or execution; visual plausibility alone cannot establish embodied realizability for the commanded action~\citep{mirabench2026}. Figure~\ref{fig:papav-anticipate} compares how forecasts are qualified and task-relevant counterfactuals selected.

\subsubsection{Anticipate: Similarity}

Digital and physical environments instantiate this function at different levels of abstraction. For multimodal agents, a candidate action may invoke a text-to-image model, predict a post-action webpage state, synthesize speech or music, generate a video, or call a software tool~\citep{gu2024your,chae2025web,fang2025webevolver}; for robotic systems, it may change future poses, velocities, contacts, collisions, object motions, rewards, and termination conditions~\citep{finn2017deep,ebert2017self,kahn2018composable,pmlr-v205-wu23c}. Nevertheless, both types of systems require a prospective account of how their available actions may change a heterogeneous multimodal environment.

\paragraph{From Current-State Representations to Predicted Environmental Change Through Predictive Models}
Anticipate provides Plan with information that cannot be obtained from the current environment state alone by modeling how the environment may change over time, across different contexts, or in response to possible external interventions. Starting from the environment state representation constructed by Perceive, both multimodal agents and robotic systems use predictive models to form probabilistic or structured expectations over possible future states. These models may take the form of learned transition models, generative observation models, latent dynamics, structured predictors, analytic or hybrid simulators, and tool or environment emulators~\citep{bruce2024genie, lin2023learning, xiang2024pandora, gu2024your, hafner2025dreamerv3, pmlr-v205-wu23c, yang2024unisim}. In multimodal agents, predictive models can capture possible changes in language-described environments, video states, webpages, GUI applications, and other interactive digital environments~\citep{gu2024your, fang2025webevolver,cao2026mobiledreamer}. In robotic systems, they can predict how control inputs may change future visual observations, object configurations, robot states, and task outcomes~\citep{ebert2017self, kahn2018composable, hu2024video}. A world model is a general form of predictive model that represents reusable environment dynamics. Compared with a conventional predictor designed to estimate a fixed target for a specific task, a world model can recursively project multiple future steps and condition those projections on different contexts or hypothetical interventions, thereby supporting a wider range of prospective queries~\citep{hafner2019dream, schrittwieser2020mastering, lin2023learning, xiang2024pandora}. Its advantage therefore lies in the generality and reusability of its dynamics representation, rather than in guaranteed predictive accuracy. Similarly, a generative model functions as a world model only when its output is treated as a possible successor state for reasoning, rather than being directly committed to the environment. The output of Anticipate is thus a predictive prior or a set of plausible future states; ranking candidate actions on the basis of those futures belongs to Plan.

\paragraph{From Predicted Change to Semantic or Latent States}
A predictor need not expose every pixel, waveform, or sensor reading in a rollout. Passing raw images, video, or multisensory trajectories downstream can be expensive and can obscure the changes that matter to the task~\citep{chae2025web,xu2026appdeltaworld,hafner2019planet,tdmpc2_2024,schrittwieser2020mastering,zhang2025up}. Semantic representations instead name expected changes in entities, events, relations, affordances, constraints, or task progress, sometimes as a delta from the current state~\citep{gu2024your,zheng2026code2world,xu2026appdeltaworld,ahn2022can}. Latent representations preserve decision-relevant dynamics without reconstructing every observation~\citep{hafner2019dream,schrittwieser2020mastering,hu2024video}. Either form may encode one future, several alternatives, or uncertainty, and hybrid representations are possible. AppDeltaWorld~\citep{xu2026appdeltaworld} predicts mobile-GUI transitions as delta code; MuZero~\citep{schrittwieser2020mastering} retains information for rewards, policies, and values in latent dynamics; and Video Prediction Policy~\citep{hu2024video} extracts predictive visual features for robot control. Across domains, Anticipate preserves the differences among possible environmental changes that matter to the next decision.

Predictive models and semantic or latent abstractions thus transform a grounded current state into task-relevant consequences of candidate interventions~\citep{xu2026appdeltaworld,hafner2019planet}. In open-ended physical interaction, plausibility alone does not make a forecast decision-worthy: simulator error, uncertain contact, and environmental change can make the same predictor useful in one regime and misleading in another~\citep{pmlr-v205-wu23c,contactworld2026,mirabench2026}. Robotic prediction already spans control-internal imagination and task-aware semantic--dynamic coupling~\citep{pmlr-v205-wu23c,zhan2026modularagent}. The remaining issue is whether a task-level decision selects the counterfactual and consumes its return. The next two contrasts address this question through comparisons with digital future prediction and control-centric robotic prediction.

\subsubsection{Anticipate: Difference}

\paragraph{MMA vs. MMEA: From Rule-Bounded Forecasts to Qualified Physical Forecasts}
Action-conditioned prediction supports candidate comparison in digital as well as embodied settings. Web Agents with World Models~\citep{chae2025web} estimates the next observation and reward for web actions and uses those estimates in policy selection; the software platform and its external services help delimit the transitions for which the model is informative, without making those transitions universally deterministic or reversible. Physical forecasts instead depend on estimated dynamics, contact, embodiment, and disturbances that remain only partly controlled at deployment. Their decision value is bounded by where and how the predictor has been evaluated. An \emph{applicability envelope} should therefore report the predicted variables and horizon, the tested operating conditions, and any signal used to detect departure from those conditions. This is a reporting requirement that links a forecast to its uncertainty and known limits, not a feature attributed to every physical predictor.

\paragraph{RS vs. MMEA: From Control-Scoped Counterfactuals to Task-Selected Counterfactuals}
Robotic world models span rather than stop at control-centric prediction. DayDreamer~\citep{pmlr-v205-wu23c} uses imagined action outcomes to improve a control policy while learning from real interaction, whereas ModularAgent~\citep{zhan2026modularagent} already couples task semantics and latent dynamics bidirectionally. ModularAgent is therefore a boundary system, not evidence that task-aware prediction is absent from robotics. PAPAV identifies the task-integrated embodied relation through a task-level route: the current decision specifies the candidate physical intervention to be queried, and the returned consequence can change commitment among candidates before execution. A modular interface may expose that route directly; for a fused policy, a prediction trace, controlled intervention, or change in candidate ranking can provide attribution evidence. No particular module boundary or model ablation is required by the definition.

\subsubsection{Anticipate: Summary}

Anticipate is present when a candidate physical intervention conditions a
prospective consequence that reaches the task decision before commitment.
Evaluation should separate errors on executed transitions and detection of
out-of-envelope cases from counterfactual measures such as ranking agreement or
decision regret. Executed outcomes are observable afterward; unchosen alternatives
require branching simulation, replay, paired trials, or repeated physical
evaluation. Representational adequacy remains task-relative. A compact state
delta can outperform a visually detailed rollout for decision support when it
preserves the differences that govern commitment, but this decision-sufficiency
criterion does not imply greater predictive accuracy across tasks or operating
regimes.

\subsection{\textcolor{plan}{P}lan}
\label{sec:plan}
\begin{figure}[!t]
  \centering
  \includegraphics[width=\linewidth,pagebox=cropbox,clip]{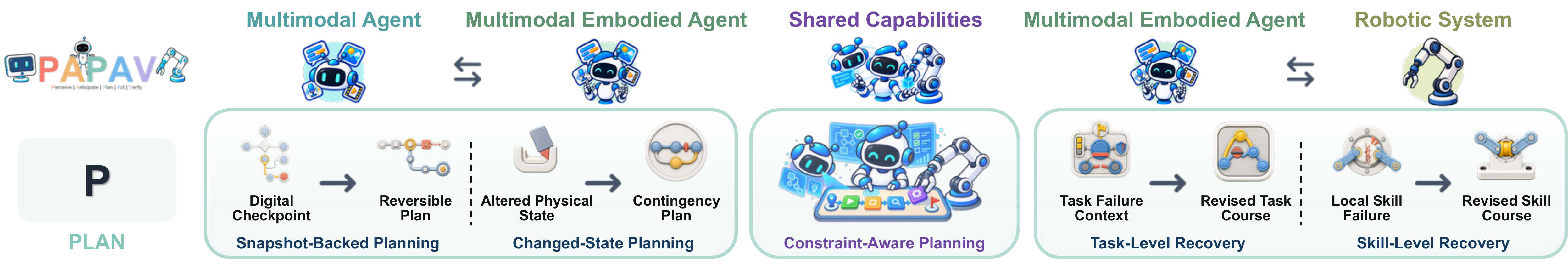}
  \caption{\textbf{Plan: MMEA vs.\ MMA vs.\ RS.}
  Shared (center): hierarchical, constraint-aware course selection. Left: checkpoint-assisted digital backtracking (a representative MMA case) vs.\ recovery planning from the reached physical state. Right: task-level committed-course revision vs.\ skill-local recovery.}
      \label{fig:papav-plan}
\end{figure}

Planning selects and organizes available or candidate actions into a goal-directed course, specifying execution order and dependencies under task constraints. Action schemas, affordances, search, and verification support this process. Language Models as Zero-Shot Planners~\citep{pmlr-v162-huang22a} maps proposed steps to admissible actions; SayCan~\citep{ahn2022can} selects skills using language scores and learned affordances; LLM-Modulo~\citep{pmlr-v235-kambhampati24a} refines candidate plans through verification. The resulting plan links selected actions to their prerequisites and subgoals. This organization determines which actions can proceed independently and which must wait for earlier steps to establish the required conditions. Physical irreversibility requires an MMEA's action selection and ordering to satisfy spatial, temporal, and safety constraints~\citep{castl2024,rekep2024,ren2023knowno}. As task conditions change, Plan can revise the remaining sequence or select an alternative course. Without a feasible course, planning may call for clarification, assistance, or abstention. Plan determines actions and their relations; Act provides body-specific execution. Figure~\ref{fig:papav-plan} compares the starting state for recovery and the decision scope for selecting a revised course.

\subsubsection{Plan: Similarity}

\paragraph{From Complex Task Goals to Hierarchically Structured Subgoals}
Planning commonly begins from a task goal, a representation of the current state, predicted consequences, and the capabilities available in the environment~\citep{gu2024your,ahn2022can,garrett2020pddlstream}. Because long-horizon tasks contain complex dependencies and combinatorial choices, both multimodal agents and robotic systems decompose goals into intermediate subgoals, order their dependencies, and progressively refine them into reusable components at finer decision scales~\citep{agashe2025agents,kaelbling2011hierarchical,sayplan2023,li2025autogui,qin2025ui}. Multimodal agents can construct workflows from GUI operations, tools, APIs, or programs, whereas robotic systems can compose navigation and manipulation stages, motion primitives, and learned skills~\citep{qin2025ui, liu2024llava,shi2025hi, liang2023code}. ProgPrompt~\citep{singh2023progprompt} expresses situated task plans as programs, while Code as Policies~\citep{liang2023code} composes perception and control APIs into executable procedures. Across digital and physical environments, the shared role of abstraction is to express the task goal and environment state against a reusable vocabulary of tools, skills, or programs, while preserving the environment-specific conditions under which each capability is available.

\paragraph{From Candidate Plans to a Selected Course under Constraints}
Once the problem is structured, Plan must deliberate over alternatives rather than continue once from the prompt. Candidate courses can be searched, scored under predicted outcomes, rejected on failed preconditions, and revised after new evidence. Multimodal agents compare webpage outcomes, choose tools, or assemble dependency-aware workflows~\citep{wu2024copilot,liu2024llava,gu2024your,shen2023hugginggpt}; robots compare skills, symbolic sequences, grasps, and trajectories under affordance, kinematic, collision, and motion constraints~\citep{garrett2020pddlstream,ahn2022can}. WebOperator~\citep{weboperator2025} ranks web actions and backtracks from unsafe paths, ControlLLM~\citep{liu2024controlllm} searches a tool graph with parameter and dependency relations, and ReKep~\citep{rekep2024} optimizes a sequence of relational keypoint constraints for manipulation. A missing API operation and an unreachable grasp thus have the same analytical effect: either can invalidate a goal-relevant candidate. Plan returns the course that survives these checks, with its intermediate objectives, ordering, dependencies, and applicability constraints~\citep{shen2023hugginggpt,liu2024controlllm,holler2021panda}.

Long-horizon decomposition and feasibility-aware selection are shared across the two system classes~\citep{shen2023hugginggpt,kaelbling2011hierarchical,liu2024controlllm,ahn2022can}. They explain coherent selection but not recovery after execution produces a discrepant state~\citep{wu2025backtrackagent,liu2023reflect,guo2024doremi}. Once a physical commitment alters the world, recoverability becomes a planning variable rather than feedback for an inexpensive retry. The contrasts below compare recovery from physical change with digital plan revision, then distinguish task-level governance of commitment from local robotic replanning during continued interaction~\citep{huang2022inner,ren2023knowno}.

\subsubsection{Plan: Difference}

\paragraph{MMA vs. MMEA: From Snapshot-Backed Planning to Changed-State Planning}
Digital and embodied agents both revise plans after execution feedback, but recovery begins from materially different states. BacktrackAgent~\citep{wu2025backtrackagent} detects GUI errors with verifier and judger modules and uses reflection to recover a productive interaction state using logged outcome pages and interaction history. This example does not make digital actions universally reversible: external services, permissions, and irreversible transactions can also remove options. A failed physical commitment may instead distribute change across geometry, contact, material condition, tool availability, and consumed resources without leaving a complete transaction record. Subsequent planning must reconstruct the state that is actually recoverable, identify effects that require compensation, and account for options removed by the preceding action. Reversibility therefore enters planning before commitment and during recovery, because it determines which remedies remain admissible under the continuing task objective.

\paragraph{RS vs. MMEA: From Skill-Level Recovery to Task-Level Recovery}
Robot planning can construct or refine a course within a specified goal and
constraint model. PDDLStream~\citep{garrett2020pddlstream} integrates symbolic search with black-box samplers, while ReKep~\citep{rekep2024} optimizes staged relational constraints for manipulation. Task-integrated embodied examples expose the return path: Inner Monologue~\citep{huang2022inner} uses scene and success feedback to select a subsequent skill, and KnowNo~\citep{ren2023knowno} makes calibrated help-seeking an admissible planning outcome. This relation does not transfer unlimited orchestration into Plan. Plan remains responsible for authorizing or revising a course under the persistent goal and constraints; Perceive updates the task belief, Verify judges the observed outcome, and a scheduler may route the resulting decision. Assistance, stopping, or recovery belongs to Plan only when it is the selected course returned to the task process. A safety controller or human may retain veto authority without becoming the planner; if the human chooses every subsequent task action, the system instead has a human-in-the-loop Plan profile.

\subsubsection{Plan: Summary}

Plan returns an authorized course under the current goal, belief, anticipated consequences, and constraints. Before commitment, the course may preserve recovery branches or choose assistance instead of physical action. After an outcome, Plan re-enters the loop only when the system selects a revised course. This boundary leaves state estimation to Perceive, outcome adjudication to Verify, and physical realization to Act. Analysis should record preconditions, protected constraints, the selected commitment, and retained recovery branches. Annotated traces can measure precondition coverage and uncompensated persistent errors; option value and expected recovery cost require branching rollouts, replay, or an explicit reachability model. Planning quality can thus reflect both the selected course and the safe, reachable alternatives it preserves, without absorbing every recovery operation into Plan.

\subsection{\textcolor{act}{A}ct}
\label{sec:act}
\begin{figure}[!t]
  \centering
  \includegraphics[width=\linewidth,pagebox=cropbox,clip]{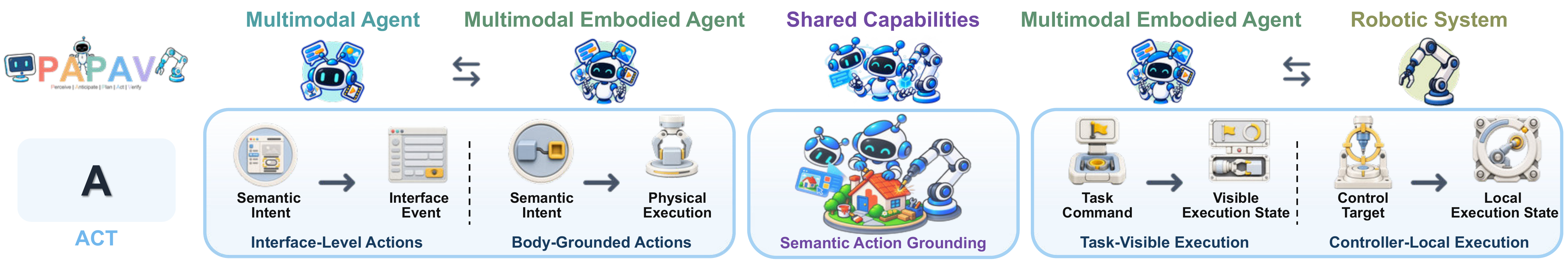}
  \caption{\textbf{Act: MMEA vs.\ MMA vs.\ RS.}
    Shared (center): semantic action grounding into environment-compatible forms. Left: compact agent-facing operations vs.\ body-grounded execution over physical time. Right: execution progress visible and consequential to the MMEA task process vs.\ controller-local execution state.}
  \label{fig:papav-act}
\end{figure}

Acting executes a selected action through the agent's interface or body. Action grounding translates intent into targets, parameters, coordinates, or control signals that the environment accepts. Closed-loop execution uses feedback to adjust commands as the action proceeds. Errors in grounding can cause even a correct plan to fail. In physical environments, execution must continue while the agent computes its next action. Asynchronous policies address this by executing one trajectory segment while generating the next. The new segment can adjust or replace the remaining motion~\citep{black2025rtc,a2c2_2025}. Execution must also account for the robot's body, since the same instruction may require different movements across robot designs. Action representations and tokenization therefore shape how a policy generates executable motion~\citep{demystifyaction2026,xtokenizer2026}. Figure~\ref{fig:papav-act} compares how actions reach the environment and how execution feedback remains available to the ongoing task process.

\subsubsection{Act: Similarity}

\paragraph{From Semantic Interventions to Grounded Action Specifications}
Plan supplies a semantic intervention; Act must still ground its target, primitive, arguments, and reference frame in the available action space. For a multimodal agent, ``click Submit'' may resolve to a GUI element and coordinate, an API call, a typed command, or code~\citep{he2024webvoyager,xu2024aguvis,zheng2024gpt}. For a robot, ``pick up the cup'' may resolve to an object, arm, coordinate frame, grasp pose, skill, or control target~\citep{jiang2023vima,oasis2026}. OS-Copilot~\citep{wu2024copilot} performs this grounding across web, terminal, file, multimedia, and application interfaces, whereas RT-2~\citep{brohan2023rt2} maps visual observations and language to robot actions. The targets differ, but the information product is the same: an environment-specific action specification accepted by the relevant interface or controller.

\paragraph{From Grounded Action Specifications to Action Representations Compatible with the Environment}
Grounding alone leaves an internal specification that must be decoded at the representation and temporal granularity required by the interface. Digital outputs include screen coordinates, structured GUI operations, commands, tool or API calls, code, and streams of discrete actions~\citep{cheng2024seeclick,gou2025navigating,hong2024cogagent,zheng2024gpt,shou2025showui,lifshitz2023steve}; robotic outputs include action tokens, chunks, waypoints, and continuous trajectories~\citep{jiang2023vima,brohan2023rt1,brohan2023rt2,zhao2023act, actionmap2026}. ShowUI~\citep{shou2025showui} emits GUI operations with interface locations, while RT-2~\citep{brohan2023rt2} generates robot actions as tokens. STEVE-1~\citep{lifshitz2023steve} produces keyboard and mouse sequences; ACT~\citep{zhao2023act} predicts short robot-action chunks; Diffusion Policy~\citep{chi2023diffusionpolicy} and RDT~\citep{liu2024rdt} generate continuous chunks through diffusion, whereas $\pi_0$~\citep{black2024pi0} uses flow matching. ShowUI-$\pi$~\citep{showuipi2026} carries this robotic recipe back into computer-use: a lightweight flow-matching action expert generates continuous cursor trajectories for drags from streaming screenshots, while discrete clicks remain in the same model, so a GUI drag is treated as a closed-loop continuous action rather than a pair of coordinates. These implementations are domain-specific, but each converts intent into executable form and can condition subsequent action on newly observed environmental feedback during execution.

The shared execution relation grounds a selected intervention in an environment-specific target and action space, encodes it at an executable temporal granularity, and permits correction as observations arrive~\citep{cheng2024seeclick,brohan2023rt2,a2c2_2025}. This explains how intent becomes command, but not how task-level meaning remains connected to body-specific control during physical execution~\citep{black2025rtc}. Relative to MMAs, a compact agent-facing operation~\citep{xie2024osworld} must expand into a two-way interface for temporally extended execution~\citep{black2025rtc}. Relative to goal-conditioned robot policies~\citep{brohan2023rt2,zhao2023act,chi2023diffusionpolicy}, body-specific control becomes task-integrated only when its execution state remains visible to, and consequential for, the ongoing task process throughout the full duration of the embodied intervention.

\subsubsection{Act: Difference}

\paragraph{MMA vs. MMEA: From Interface-Level Actions to Body-Grounded Actions}
MMA environments can expose relatively compact agent-facing operations. OSWorld~\citep{xie2024osworld}, for example, presents clicks, text entry, code, and other platform-defined interfaces to the agent. This compactness is a property of the interface rather than of the setting: dragging a slider or a timeline already demands on-the-fly perception and adjustment, and ShowUI-$\pi$~\citep{showuipi2026} meets it with the action expert borrowed from robot policies~\citep{black2024pi0}. Physical execution extends the same requirement across embodiment and time: a particular body must realize the semantic commitment while the task process retains access to its progress. A bidirectional execution contract makes this relation observable. Its downward record identifies the capability, target, parameters, preconditions, and admissible interruption points; its upward record may include controller state, estimated state change, progress, termination status, low-level faults, and observation time. The two directions keep the task meaning of the commitment traceable while leaving implementation and timing to embodiment-specific control. These fields are proposed observability criteria, not a claim that current MMEAs share a standardized execution API.

\paragraph{RS vs. MMEA: From Controller-Local Execution to Task-Visible Execution}
Robotic policies already realize compact objectives on particular bodies. RT-2~\citep{brohan2023rt2}
tokenizes robot actions, whereas ACT~\citep{zhao2023act} and Diffusion Policy~\citep{chi2023diffusionpolicy} produce temporally
extended motor sequences. Within PAPAV,
these systems establish body-specific execution but not, by that fact alone, a
task-integrated embodied Act relation. Attribution depends on whether the current
task process retains the authorized semantic commitment and can use time-sensitive
execution feedback while intervention remains useful. Coding should check whether the downward record identifies the target,
intended state change, governing constraints, and interruption conditions, and
whether the upward record exposes progress, termination, faults, and timing. Such telemetry
neither substitutes for Perceive's estimate of the physical state nor determines
Verify's postcondition verdict. An explicit API, a modular control stack, or an
end-to-end policy can instantiate Act, provided that time-sensitive
execution feedback can still modify the course.

\subsubsection{Act: Summary}

Act maintains continuity between an authorized intervention and its body-specific realization. Evaluation should measure grounding correctness, progress observability, termination-state accuracy, interruption latency, and missed deadlines when the interface exposes the necessary evidence. Semantic divergence should compare the authorized target, intended state change, and protected constraints with the execution record rather than compare two free-form explanations. These measures diagnose Act directly; the final task verdict remains Verify's responsibility. Specialized control is compatible with task-level agency when the authorized commitment remains traceable and timely feedback can still change the course. Because the grounding-and-decoding relation is shared, action machinery also crosses the boundary: ShowUI-$\pi$~\citep{showuipi2026} transfers flow-based action generation from robot-use to GUI dragging, illustrating a shared Act mechanism.

% ======== Verify section modified by ziyi for adding self-improvement loop
\subsection{\textcolor{verify}{V}erify}
\label{sec:verify}
\begin{figure}[!t]
  \centering
  \includegraphics[width=\linewidth,pagebox=cropbox,clip]{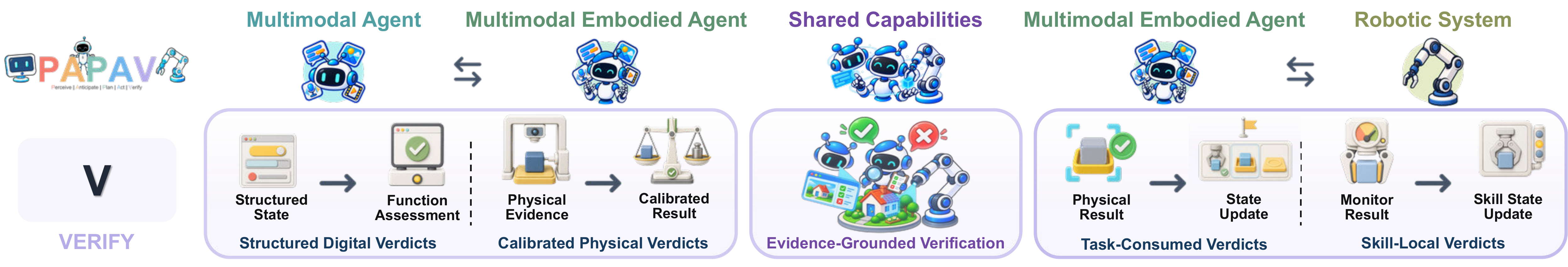}
  \caption{\textbf{Verify: MMEA vs.\ MMA vs.\ RS.}
  Shared (center): evidence-grounded verification against independently specified expectations. Left: structured digital checks vs.\ physical verdicts calibrated under partial, uncertain, potentially delayed evidence. Right: task-consumed vs.\ locally used monitor or skill verdicts.}
  \label{fig:papav-verify}
\end{figure}

Verification judges intended change, task progress, and constraint satisfaction from post-action evidence~\citep{kolthoff2025guispector,vagen2026,shi2026interactive}. MMAs and RSs compare completion criteria with observed consequences, including GUI states, application data, camera observations, proprioception, contact signals, and execution traces. WebArena~\citep{zhou2024webarena} and OSWorld~\citep{xie2024osworld} check page, application, or execution states; Inner Monologue~\citep{huang2022inner} returns scene and success signals to planning; vision-language success detectors~\citep{du2023successdetectors} infer outcomes from observations. Embodied verification remains retrospective, but physical interaction often lacks an independent, fully observable success signal. Perceive maintains a task-general belief; Verify tests goal-specific criteria against post-action evidence. Although learned evaluators support action selection~\citep{robomonkey2025} and progress rewards for learning~\citep{progressrewardsurvey2026}, within PAPAV, Verify assesses executed actions, supporting task correction and learning. Figure~\ref{fig:papav-verify} distinguishes the evidence supporting an outcome judgment from the level at which that judgment is consumed.

\subsubsection{Verify: Similarity}

\paragraph{From Observed and Expected Outcomes to Verification Judgments}
Verification requires two grounded inputs: evidence of what occurred after an action and an independently specified account of what should have occurred~\citep{ramrakhya2026scaling,luo2024vision}. In digital environments, such evidence may include tool returns, outcome pages, persistent interface states, and interaction histories; in physical environments, it may additionally include visual, proprioceptive, contact, force, and other temporally accumulated observations. These signals allow GUI agents to inspect state transitions and robots to reconstruct manipulation outcomes~\citep{liu2023reflect,wu2025backtrackagent,guo2026agent,gao2026guitester}. The corresponding reference may be a task goal, predicted next state, subgoal condition, or explicit success criterion. Self-Grounded Verification~\citep{andrade2026let} addresses this problem by first eliciting task-completion criteria independently of the candidate trajectory and only then evaluating the trajectory against those criteria; importantly, the same mechanism is demonstrated across web agents, GUI agents, and robotic manipulation policies. Physical-domain critics additionally require causal, spatial, and affordance awareness, while runtime failure detectors must account for uncertainty and out-of-distribution execution states~\citep{xiong2026phycritic,xu2025can}. Earlier reflection-based methods establish useful feedback--revision loops, but self-generated critique alone does not guarantee correspondence with the actual environment~\citep{shinn2023reflexion,madaan2023self}. Verification should therefore be grounded in independently specified criteria, multimodal observations, tool-accessible states, or calibrated anomaly evidence.

\paragraph{From Verification Judgments to Corrective Feedback}
Verification must convert the gap between expected and observed outcomes into feedback that regulates the loop, not merely a detached binary label. The verdict may distinguish terminal success, intermediate progress, recoverable mismatch, insufficient evidence, an invalid expectation, or persistent risk. Language-conditioned success classifiers and multimodal reward models turn observed state changes into execution signals that can also support policy learning~\citep{nair2022learning,wu2021example,ma2023liv,wang2024rl,luu2025enhancing, lin2026paperdoctor}. In digital environments, application state or a failed tool return can reveal task progress; Voyager~\citep{wang2024voyager} combines environment feedback, execution errors, and self-verification to revise later behavior. Physical verification can instead depend on object relations, poses, or contact. REFLECT~\citep{liu2023reflect} summarizes robot experience to diagnose failures, and DoReMi~\citep{guo2024doremi} recovers from plan--execution mismatch. An actionable verdict should locate the mismatch and indicate the corrective response, rather than report correctness alone.

Corrective feedback should distinguish outcome assessment from failure attribution. After a placement attempt, an object outside the target region indicates an unmet postcondition, but does not establish whether the cause was incorrect grounding, an unsuitable plan, or execution error. The verifier should report the failed condition and supporting evidence. This distinction helps the task process select an appropriate response without treating every failure as a planning error.

\paragraph{From Outcome Verification to Policy Improvement}
Verify can support improvement beyond the current task. In a self-improvement loop, the agent uses evaluated experience to update its policy. The updated policy then generates further attempts and new evidence for verification. Verify supplies evidence for learning; the learning mechanism performs the update.

VERITAS reuses successful trajectories collected under verifier guidance for policy training~\citep{veritas2026}. VLAC and EVOLVE-VLA use learned progress feedback to support policy updates through interaction~\citep{vlac2025,evolvevla2025}. These methods connect evaluation signals to policy improvement. A successful retry alone does not demonstrate a lasting improvement. The claim requires a retained update and better performance in subsequent evaluation. Incorrect judgments can admit failed behavior into training data or reward unhelpful actions.

Verification closes the loop by comparing post-action evidence with independently specified expectations~\citep{andrade2026let,luo2024vision} and returning a judgment that can regulate later execution~\citep{liu2023reflect}. Physical evidence may be partial or delayed~\citep{sentinel2024,checkvla2026}, observations uncertain~\citep{xu2025can}, and the verifier dependent on the same evidence as the acting stack. The MMA comparison therefore moves from software-mediated functional checks~\citep{zhou2024webarena,li2026mobileuse} to physical verdicts whose timing and dependencies shape reliability~\citep{xiong2026phycritic}. The robotics comparison asks whether an outcome signal remains local to monitoring or reaches the task process in time to influence the same episode~\citep{du2023successdetectors,guo2024doremi}.

\subsubsection{Verify: Difference}

\paragraph{MMA vs. MMEA: From Structured Digital Checks to Evidence-Grounded Physical Verification}
Digital evaluation must be separated from agent-side verification. WebArena~\citep{zhou2024webarena} uses external programs to judge functional correctness, whereas MobileUse~\citep{li2026mobileuse} places hierarchical reflection and recovery inside the agent's ongoing task loop. Only the latter establishes an agent-consumed Verify relation. In an MMEA, the judgment rests on post-action physical evidence affected by partial visibility, sensor uncertainty, and delayed consequences. Timeliness becomes part of correctness: a verdict must reach the task process while continuation, renewed perception, replanning, recovery, assistance, or stopping can still change the outcome. Analysis should therefore record the verifier's evidential-dependency profile. Shared sensors, representations, model backbones, or training data may produce correlated errors despite architectural separation, so verification records should disclose provenance and dependencies and quantify correlated failures wherever the evaluation design supports comparison against the policy that produced the action under evaluation. Uncertain verdicts may become unreliable training targets. Repeated updates can reinforce errors shared by the actor and verifier. Later evaluation should therefore check improvement against observed task outcomes.

\paragraph{RS vs. MMEA: From Skill-Local Outcome Detection to Task-Consumed Verification}
Robotic outcome detectors can serve training, offline evaluation, local control, or the ongoing task process. Vision-Language Models as Success Detectors~\citep{du2023successdetectors} turns observed behavior into a reward or evaluation signal; this output alone does not show that a task-level agent consumes the verdict during the episode. DoReMi~\citep{guo2024doremi}, which appears here as task-integrated embodied evidence, instead detects plan--execution mismatch and routes the result into corrective replanning. Verify acquires its task-integrated embodied role when an actionable verdict reaches the current task process before selection of the next course. Coding should record the judged outcome, confidence, supporting evidence, unresolved conditions, consumer, and intervention window. A controller stop, a task-level verdict, and an offline label may share a detector while retaining different responsibilities.

\subsubsection{Verify: Summary}

Verify adjudicates whether post-action evidence satisfies task postconditions and returns a verdict that can change the current episode, including termination. Act may provide execution telemetry and low-level faults; Verify assigns success, failure, partial progress, or unresolved uncertainty against task criteria. Annotated episodes support direct measurement of false acceptance, false rejection, calibration error, and verdict latency. Intervention benefit and recovery utility are causal quantities that require matched runs, randomized verifier use, branching replay, or an explicit causal estimator. Reports should pair verdict accuracy with evidence sources, dependencies on the acting system, and correlated errors. They should also identify who uses the verdict and when intervention remains possible. For self-improvement, evaluation should test whether the retained update improves performance in further trials under comparable conditions.

\subsection{Synthesis: From Capability Evidence to Evaluation}
\label{sec:papav-synthesis}
PAPAV capabilities serve the same core functions across digital and physical systems. Physical embodiment changes how these functions are carried out. The comparison with MMAs shows how physical interaction changes multimodal sensing, prediction, planning, execution, and outcome verification. The comparison with RSs examines how these capabilities inform ongoing task decisions. The analysis considers individual capabilities and how their outputs guide subsequent behavior.

Studies spanning digital and physical environments provide evidence for shared capabilities~\citep{szot2025multimodal}. Recent work also examines capability coordination, failure diagnosis, and interactive system evaluation~\citep{xu2026roboagent,ying2026agentsafe,cheng2026embodiedeval}. These findings support capability analysis. Classification of complete systems follows the conditions in Section~\ref{sec:bg-mea}.

The PAPAV framework guides the analysis of benchmark coverage. For each capability, evaluation should identify its inputs, outputs, and role in task decisions. It should also specify whether the capability is explicitly assessed. Further analysis should examine how uncertainty, execution feedback, and recovery affect later decisions. Some systems continue to learn through interaction. Their evaluation should identify what is updated and whether these changes persist over repeated interactions. Further trials should test their effects on task performance under comparable conditions. A successful retry alone does not demonstrate lasting improvement. Section~\ref{sec:benchmark} examines how existing benchmarks assess the five capabilities and where gaps in evaluation remain.

%% file: sections/4-benchmark.tex
\section{Benchmarks \& Evaluation}
\label{sec:benchmark}
This section compares how existing benchmarks evaluate PAPAV capabilities across multimodal embodied agents (MMEAs), multimodal agents (MMAs), and robotic systems (RSs). We distinguish task requirements from explicit evaluation:a task may require a capability, but overall outcome metrics do not reveal how well the agent performs that capability. Our curated corpus contains 64 benchmarks: 10 MMEA, 23 MMA, and 31 RS benchmarks. Table~\ref{tab:benchmark-main-table} summarizes their information, task settings, and PAPAV coverage. All counts and proportions refer to this corpus unless stated otherwise.
\paragraph{Benchmark Collection and Selection}
Candidates were identified through Google Scholar and relevant publication venues. Inclusion required defined tasks or environments, a public protocol with at least one quantitative metric, PAPAV relevance, and sufficient evidence for capability labeling. Successor benchmarks were counted separately only if tasks, environments, datasets, or protocols changed substantially. Benchmarks were assigned to mutually exclusive MMEA, MMA, and RS categories by their primary evaluation domains; multimodal input alone did not qualify an RS benchmark as MMEA. The corpus was last updated on 8 Sept 2026.

\paragraph{Capability Labeling Protocol}
Each of the curated benchmarks was assessed across the five PAPAV capabilities. The labels have three levels:
\begin{itemize}
    \item \textbf{Explicit Evaluation (\directcov)}: A dedicated task, metric, controlled comparison, or observable signal provides evidence specific to the capability.
    \item \textbf{Implicit Evaluation (\partialcov)}: The capability is required but appears only in end-to-end performance.
    \item \textbf{Not Evaluated (\nonecov)}: The capability is not required by the task or evaluation protocol.
\end{itemize}

The following capability boundaries prevent overestimation:
\begin{enumerate}
    \item \textbf{Perceive}: Perceptual understanding, grounding, or information extraction must be assessed separately. Multimodal input alone is insufficient.
    \item \textbf{Anticipate}: Explicit Evaluation of Anticipate requires scored predictions of action consequences, risks, or counterfactual states. Ordinary state changes after actions are insufficient.
    \item \textbf{Plan}: Task-level planning, skill organization, or replanning must be assessed. Path planning, motion planning, trajectory generation, and state-space search are not task-level Plan.
    \item \textbf{Act}: Action execution, validity, or environmental effects must be assessed. Producing action descriptions or future videos is insufficient.
    \item \textbf{Verify}: The agent must judge success or failure from feedback. Checks performed only by external evaluators do not assess Verify.
\end{enumerate}

% ------------------------ START of Benchmark Table -----------------------
\input{tables/Benchmark_Table}
% ------------------------ END of Benchmark Table ------------------------

% -------------------MMEA
\subsection{Multimodal Embodied Agents}
\label{sec:MMEA}
Because multimodal embodied tasks often involve several PAPAV capabilities simultaneously, we distinguish capability involvement from Explicit Evaluation, which requires capability-specific evidence but does not necessarily isolate every possible source of failure.

MMEA benchmarks connect vision--language reasoning with physical action. All ten explicitly evaluate Act, and most explicitly evaluate Perceive and Plan. Anticipate is always Implicit; only RoboCerebra~\citep{han2026robocerebra} and AgenticLab~\citep{guo2026planar} explicitly evaluate Verify. Current protocols cover the main ``Perception--Planning--Action'' chain but provide weaker diagnosis of predicted consequences and outcome verification. Most rely on simulation, with limited real-world and sim-to-real evaluation.

\subsubsection{Embodied Perception-Action}
IS-Bench~\citep{lu2026bench} evaluates safety risks that dynamically emerge during household tasks. ESI-Bench~\citep{hong2026esi} requires agents to actively acquire spatial information relevant to the task through movement and interaction. CaP-Bench~\citep{fu2026cap} evaluates executable robot control code generated from visual observations and language instructions. Agents can iteratively refine the code using environmental feedback.

All three benchmarks explicitly evaluate Perceive and Act through embodied interaction. Anticipate remains Implicit because predictions of future states are not independently scored. Plan is Implicit in IS-Bench and CaP-Bench because task organization is reflected only in the final outcome. ESI-Bench explicitly evaluates task-level Plan through the selection and sequence of perception, locomotion, and manipulation behaviors. IS-Bench and ESI-Bench do not evaluate Verify. Their agents are not required to judge action success and respond accordingly. Verify is Implicit in CaP-Bench through iterative code refinement based on environmental feedback.

\subsubsection{Planning and Closed-Loop}
TEACh~\citep{padmakumar2022teach} evaluates dialogue-guided household tasks. EMBODIEDBENCH~\citep{yang2025embodiedbench} covers household interaction, navigation, manipulation, perception, and long-horizon planning. RoboCerebra~\citep{han2026robocerebra} uses a hierarchical architecture with planning, control, memory, and verification. AgenticLab~\citep{guo2026planar} evaluates task planning, motion checking, goal verification, and replanning under real-world perturbations.

TEACh explicitly evaluates Act. Its other four capabilities are Implicit. EMBODIEDBENCH explicitly evaluates Perceive, Plan, and Act. Anticipate and Verify are Implicit. RoboCerebra and AgenticLab explicitly evaluate Plan, Act, and Verify. Perceive is Explicit in AgenticLab and Implicit in RoboCerebra. Anticipate is Implicit in both.

\subsubsection{Multi-Agent Coordination}
PARTNR~\citep{chang2025partnr} evaluates human-robot collaboration under spatial, temporal, and capability constraints. RoCoBench~\citep{mandi2024roco} evaluates collaborative strategies, task assignment, and execution planning. Habitat-MAS~\citep{chen2025emos} evaluates task allocation and coordination among heterogeneous robots.

All three explicitly evaluate Plan and Act. PARTNR and Habitat-MAS also explicitly evaluate Perceive through environmental understanding and capability grounding. RoCoBench uses textual or structured states and does not evaluate Perceive. Anticipate is Implicit because coordination depends on other agents' actions and constraints. Verify is Implicit through feedback, task progress, and recovery. External collision and simulator checks do not assess verification by the agent.

% -------------------MM Agents
\subsection{Multimodal Agents}
\label{sec:MMA}
MMA benchmarks range from multimodal understanding to interactive agents, artifact generation, and world models. Understanding benchmarks primarily evaluate Perceive~\citep{yue2025mmmu,yang2025thinkinginspace,fu2026video}. Web, computer, mobile, and game benchmarks explicitly evaluate Act and often cover Plan and Anticipate implicitly~\citep{zhou2024webarena,yuan2026osworld2,rawles2025androidworld,ouyang2026gameworld}. Artifact benchmarks treat executable outputs as Act~\citep{luo2026gamecraft,jimenez2024swe,lu2026webgen}, while world model benchmarks explicitly evaluate Anticipate~\citep{nvidia2025pbench,li2026worldmodelbench,mirros2026harnessevalw}. Verify remains limited despite the post-action feedback available in several benchmarks~\citep{mayer2025ivispar, ouyang2026gameworld, tian2025mmina, tao2026mmsearch}, because such feedback rarely requires the agent to judge success and decide whether to stop, retry, or recover. Checks performed only by the evaluator do not count as Verify.

% \vspace{-3mm}
\subsubsection{Multimodal Understanding}
MMMU-Pro~\citep{yue2025mmmu}, Video-MME-v2~\citep{fu2026video}, and VSI-Bench~\citep{yang2025thinkinginspace} target interdisciplinary multimodal question answering, video understanding, and visual-spatial reasoning, respectively. MMMU-Pro reduces the influence of linguistic priors on results by filtering out questions that can be answered from text information alone. Video-MME-v2 evaluates video understanding through visual information aggregation, temporal dynamic modeling, and cross-temporal reasoning. VSI-Bench evaluates spatial relations, distance, orientation, and scene memory based on egocentric indoor video. These benchmarks provide capability-specific evaluation rather than a complete agent loop evaluation.

\subsubsection{Web Browsing, Computer-Use, and Mobile Interaction}
WebArena~\citep{zhou2024webarena}, VisualWebArena~\citep{koh2024visualwebarena}, WebVoyager~\citep{he2024webvoyager}, MMInA~\citep{tian2025mmina}, and MMSearch-Plus~\citep{tao2026mmsearch} evaluate web browsing and search. All five explicitly evaluate Act and implicitly evaluate Plan. WebArena uses structured webpage representations and does not evaluate visual perception. VisualWebArena, MMInA, and MMSearch-Plus explicitly evaluate Perceive. Perceive is Implicit in WebVoyager. Anticipate is Implicit in all except MMSearch-Plus. WebArena, VisualWebArena, and WebVoyager implicitly evaluate Verify through action result inspection. MMInA and MMSearch-Plus do not evaluate Verify because their checks are external.

AndroidWorld~\citep{rawles2025androidworld}, OSWorld 2.0~\citep{yuan2026osworld2}, and AgentStudio~\citep{zheng2025agentstudio} extend the evaluation to mobile devices, desktop systems, and general virtual environments. AndroidWorld and OSWorld 2.0 explicitly evaluate Act. Perceive, Anticipate, Plan, and Verify are Implicit through long-horizon interaction with dynamic interfaces. AgentStudio explicitly evaluates Perceive, Act, and Verify. AgentStudio covers Anticipate and Plan only implicitly.

\subsubsection{General and Cross-Environment Assistants}
GAIA~\citep{mialon2024gaia} evaluates document understanding, web search, reasoning, and tool use through real-world questions with deterministic answers. CRAB~\citep{xu2025crab} covers workflows across desktop and mobile environments. AgentVista~\citep{su2026agentvista} focuses on hybrid tool use guided by visual evidence. OmniGAIA~\citep{li2026omnigaia} evaluates multimodal reasoning across images, video, and audio.

GAIA evaluates Perceive, Plan, and Act implicitly through final answer accuracy. It does not evaluate Anticipate or Verify. CRAB explicitly evaluates Act through graph-based progress checks. Perceive, Anticipate, Plan, and Verify are Implicit. AgentVista and OmniGAIA explicitly evaluate Perceive and Act. Plan is Implicit through multi-step tool use. Anticipate and Verify are Not Evaluated. Their tools support information acquisition, and their outputs are checked externally.

\subsubsection{Interactive Games and Spatially Intelligent Agents}
GameWorld~\citep{ouyang2026gameworld} evaluates visual interaction across multiple browser-based games. Agents continuously process visual feedback and execute keyboard, mouse, or semantic actions in the game environment. iVISPAR~\citep{mayer2025ivispar} uses interactive sliding puzzles to evaluate spatial reasoning and sequential manipulation. It compares agent performance across visual and textual state representations.

Both benchmarks explicitly evaluate Act and implicitly evaluate Anticipate. iVISPAR explicitly evaluates Perceive through controlled representation comparisons. Perceive is Implicit in GameWorld. GameWorld evaluates Plan implicitly through multi-stage objectives. Local moves and state space search in iVISPAR are not task-level Plan. Neither benchmark evaluates Verify. State observation is part of normal interaction, and evaluator checks do not require agent judgment or recovery.

\subsubsection{Artifact Generation}
SWE-bench~\citep{jimenez2024swe}, WebGen-Bench~\citep{lu2026webgen}, and GameCraft-Bench~\citep{luo2026gamecraft} evaluate software issue resolution, interactive website generation, and playable game generation, respectively. In these tasks, Act extends to generating artifacts that can be compiled, run, or interactively tested.

Perceive and Plan are Implicit because they affect artifact quality but are not evaluated separately. Anticipate and Verify are Not Evaluated because artifact testing is external and does not require prediction, failure detection, or revision.

\subsubsection{World Model and Physical Prediction}
WorldModelBench~\citep{li2026worldmodelbench} and PBench~\citep{nvidia2025pbench} evaluate future video generation from initial visual states and textual inputs, focusing on temporal consistency, physical plausibility, and condition compliance. HarnessEval-W~\citep{mirros2026harnessevalw} evaluates interactive world predictions through observation quality, transition correctness, and world persistence. Its hierarchical evaluator decomposes each case into measurable subproblems and aggregates validated evidence into a traceable score.

All three explicitly evaluate Anticipate through scored future predictions. Perceive is Implicit because understanding the initial scene is required but not assessed separately. Generated observation quality does not independently establish perceptual understanding. Act is Not Evaluated because the model outputs are predicted observations. The protocols do not assess task-level planning or feedback-driven verification by the evaluated models, so Plan and Verify are also Not Evaluated. In HarnessEval-W, planning and evidence validation belong to the external evaluator.

% -------------------Robotic Systems
\subsection{Robotic Systems}
\label{sec:RS}
The evaluation focus of RS benchmarks is concentrated. Of the 31 benchmarks in Table~\ref{tab:benchmark-main-table}, 29 provide Explicit Act evidence through task outcomes, return, path efficiency, or stage completion. Aggregate scores combine execution errors with perception or planning errors. None explicitly evaluates task-level Plan or the agent's Verify capability. Path and motion planning, trajectory optimization, and low-level control count as execution unless goal decomposition, skill sequencing, or replanning is assessed. Simulation, physical deployment, and hybrid testing are evaluation settings.

\subsubsection{Navigation and Exploration}
R2R~\citep{anderson2018vision} evaluates instruction following from egocentric vision using endpoint success and path efficiency. GOAT-Bench~\citep{khanna2024goat} uses sequences of category, language, and image goals, making spatial memory and information reuse part of the task. Both provide Explicit Perceive and Act evidence through visual grounding and navigation. Their aggregate scores still combine target recognition, localization, memory, and movement errors.

Anticipate and Verify are Implicit because state prediction and progress checking affect successful behavior without separate scores. Path search is not treated as task-level Plan. Memory-specific errors and goal-level trajectories would make the different episode structures easier to compare.

\subsubsection{Skill Learning and Policy Generalization}
RLBench~\citep{james2020rlbench}, ManiSkill2~\citep{gu2023maniskill2}, RoboSuite~\citep{zhu2020robosuite}, and Meta-World+~\citep{mclean2026meta} provide common manipulation tasks, robot models, controllers, and skill collections. Their main success and return metrics make Act observable but do not isolate individual failure causes. LIBERO~\citep{liu2023libero} evaluates new skill acquisition and retention. LIBERO-PRO~\citep{zhou2025libero} adds visual and task perturbations. LIBERO-Plus~\citep{fei2025libero} evaluates manipulation robustness across seven perturbation dimensions with tasks at five difficulty levels. CALVIN~\citep{mees2022calvin} evaluates language-conditioned skill sequences. RoboCasa365~\citep{nasiriany2026robocasa365} extends these sequences to varied kitchen tasks. RoboLab~\citep{yang2026robolab} introduces controlled task and environment changes. FMB~\citep{luo2025fmb} evaluates individual and composed skills on a reproducible physical platform. These benchmarks cover retention, perturbation robustness, skill composition, and physical variation.

CALVIN, LIBERO-PRO, LIBERO-Plus and RoboLab explicitly evaluate Perceive. Perceive is Implicit in most other suites. Meta-World+ uses structured states and does not evaluate Perceive. Plan is Implicit in LIBERO, LIBERO-PRO, LIBERO-Plus, RoboCasa365, RoboLab, and FMB. It is Not Evaluated in RLBench, ManiSkill2, RoboSuite, Meta-World+, and CALVIN. None evaluates Verify because completion checks are external.

\subsubsection{Long-Horizon and Household Manipulation}
ALFRED and VLABench connect language, navigation, and manipulation in multi-stage embodied tasks. A valid individual action can still fail when performed at the wrong point in the task sequence~\citep{shridhar2020alfred,zhang2025vlabench}. BEHAVIOR-1K and HomeRobot expand the task space to object search, mobile manipulation, and diverse everyday activities~\citep{li2023behavior,yenamandra2023homerobot}. FurnitureBench and the BEHAVIOR Robot Suite place long-horizon assembly and household tasks on real physical platforms, adding contact errors, hardware tolerances, and reset variation~\citep{heo2023furniturebench,jiang2025behavior}.

Reported goal conditions and phase-level progress can help distinguish early failures from near completion. These measures can still conflate semantic, perceptual, navigation, and manipulation errors. Progress is also computed by an external evaluator. It does not necessarily show how the robot organized the task or recognized a failed intermediate state. These aggregate measures do not independently establish task planning or agent verification.

\subsubsection{Safety-Constrained Control and Human Assistance}
Safety-Gymnasium~\citep{ji2023safety} reports task reward and constraint cost. SafeManip~\citep{huang2026safemanip} checks prohibited states and temporal-ordering violations. NIABench~\citep{zhang2026assistance} evaluates the utility and timing of assistance, including cases where inaction is preferable.

These protocols evaluate whether actions are safe, extending Act beyond task completion. Anticipate is Implicit because risk affects behavior without a separate score for predicted consequences. Plan is Implicit only in SafeManip, whose ordered constraints provide evidence of task organization. Constraint monitoring and intervention scoring remain external, so none evaluates Verify performed by the agent.

\subsubsection{Real-World and Sim-to-Real Evaluation}
RoboArena~\citep{roboarena2025} distributes pairwise physical-policy comparisons across participating sites, reducing dependence on the task setup of a single laboratory. VLA-REPLICA~\citep{huang2026vla} specifies a lower-cost hardware configuration for repeatable in- and out-of-distribution tests. RoboChallenge~\citep{robochallenge2025} organizes centralized evaluation at a larger scale. RoboTwin~\citep{mu2025robotwin} and RoboTwin~2.0~\citep{chen2025robotwin} use digital twins to expand scene and bimanual variation before deployment. RobotArena$\infty$~\citep{jangir2025robotarena} asks whether policy rankings in reconstructed simulations agree with physical trials. RoboDojo~\citep{robodojo2026} connects simulation with standardized remote robot experiments.

These benchmarks focus on policy performance across sites, hardware platforms, and sim-to-real settings. All explicitly evaluate Act through executed robot behavior. Perceive is Explicit only when a protocol controls and compares visual conditions. Plan and Verify are Not Evaluated because task generation, state checking, and ranking are handled by the evaluation system rather than the policy. Ranking agreement measures sim-to-real validity. Site-level results reveal variations caused by cameras, control frequency, reset procedures, or operators.

\subsubsection{Robot World Models and Physical Prediction}
Dream.exe~\citep{zhao2026dream} evaluates generated manipulation futures through video quality, trajectory consistency, and downstream execution. RoboWM-Bench~\citep{jiang2026robowm} evaluates whether predicted futures are physically plausible and useful for manipulation. Both explicitly evaluate Anticipate. Perceive is Implicit because the initial image is only an input condition. Downstream execution provides indirect evidence for Act.

The evaluated models stop after generating a useful future prediction. They do not complete a full decision loop, which would compare alternative futures, select an action, observe the outcome, and update the next prediction. Plan and Verify are therefore Not Evaluated. A separate control policy remains responsible for action execution. Recording the selected prediction and its later update would better connect Anticipate with closed-loop behavior.

\subsection{Cross-Domain Comparison}
\begin{figure}[t]
    \centering
    \begin{minipage}[c]{0.548\linewidth}
        \centering
        \includegraphics[width=\linewidth]
        {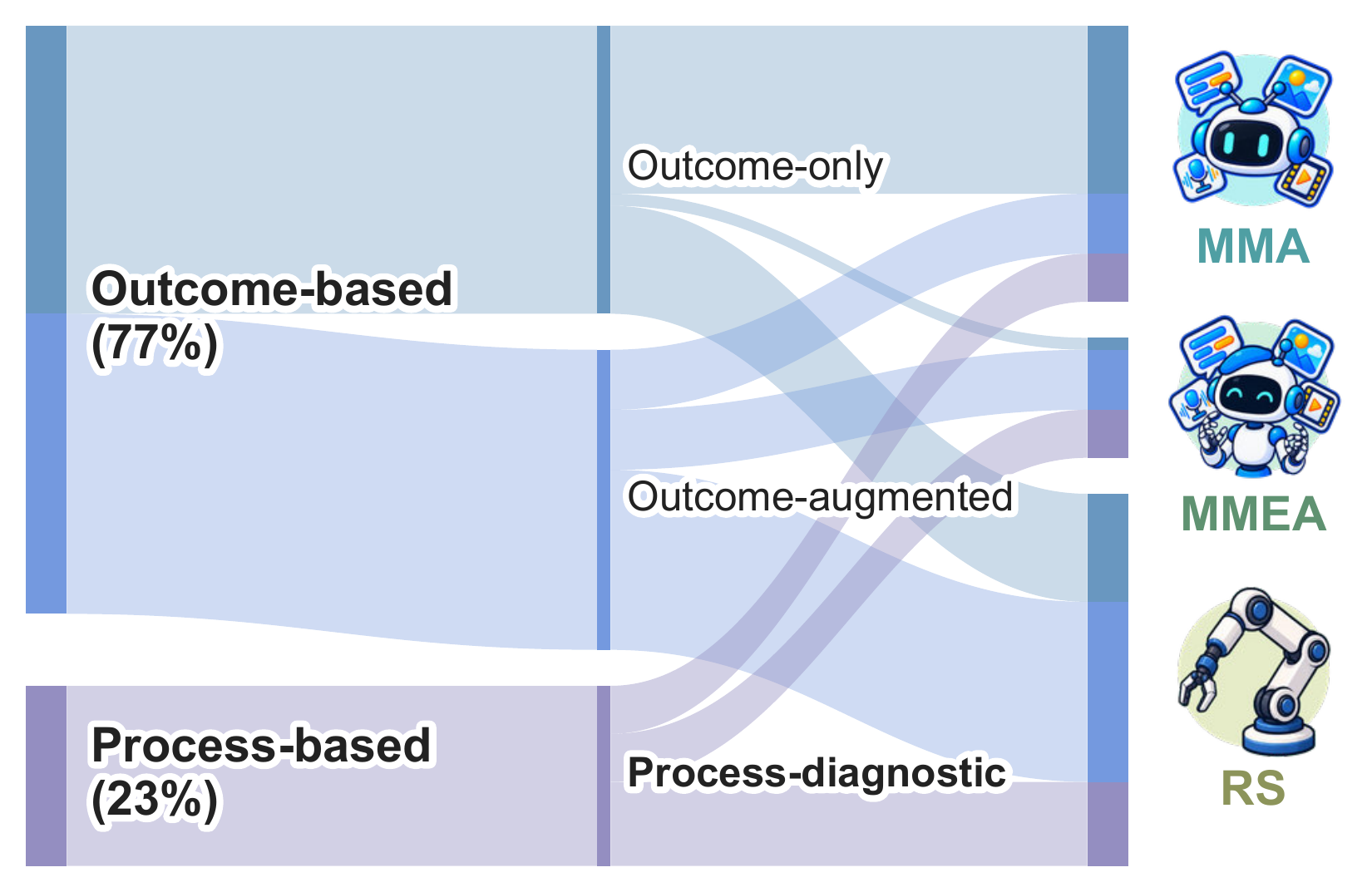}
    \end{minipage}\hfill%
    \begin{minipage}[c]{0.448\linewidth}
        \centering
        \includegraphics[width=\linewidth]
        {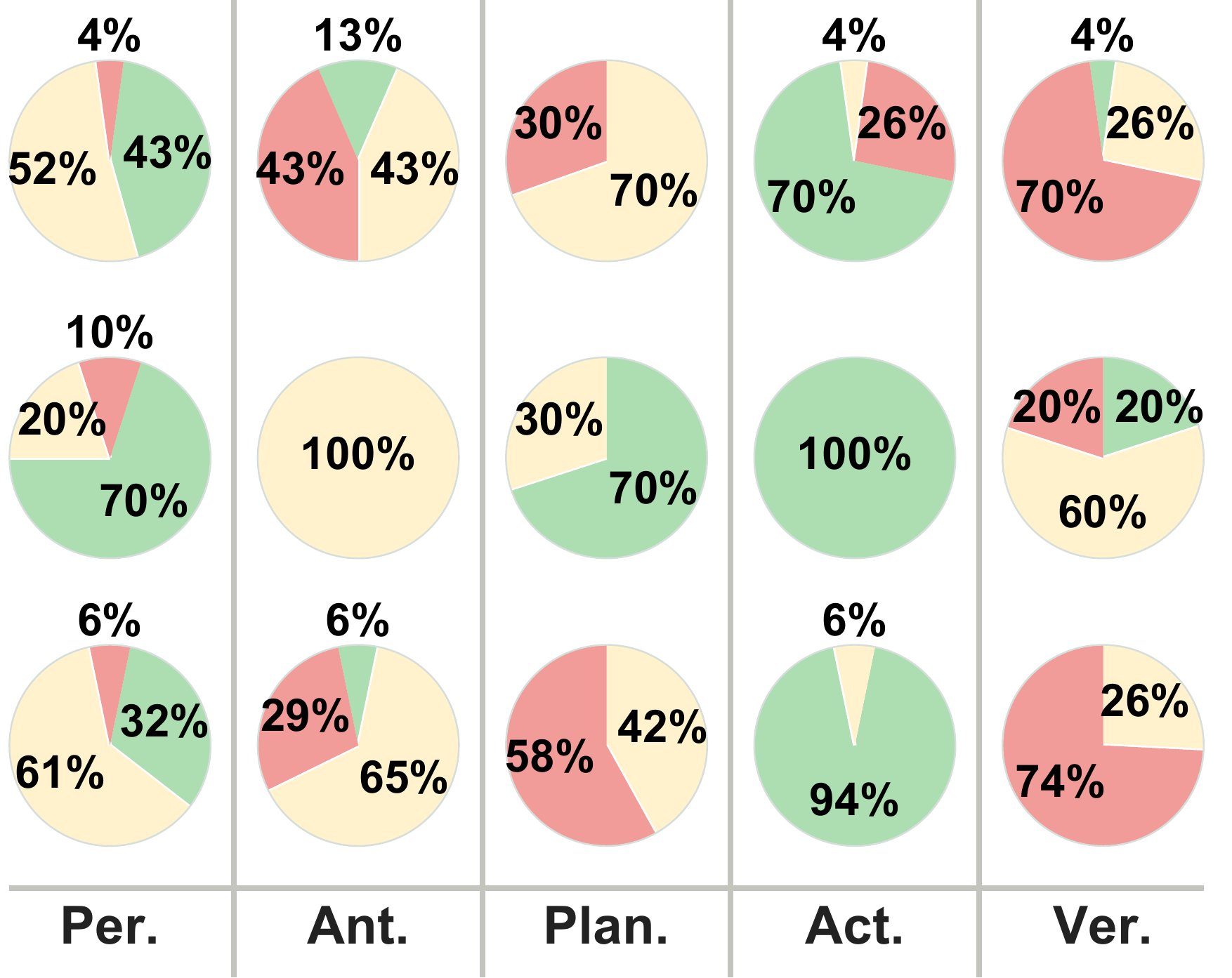}
    \end{minipage}

    \caption{\textbf{Analysis of benchmark metrics and cross-domain PAPAV coverage.}
    Left: Sankey diagram tracing benchmark metric types across MMA, MMEA, and RS benchmarks. Right: Three-level PAPAV capability coverage matrices for MMA, MMEA, and RS benchmarks. Green, yellow, and red denote Explicit Evaluation, Implicit Evaluation, and Not Evaluated, respectively.}
    \label{fig:benchmark-analysis}
\end{figure}

Figure~\ref{fig:benchmark-analysis} summarizes the 64 retained benchmarks. The Sankey diagram groups 49 outcome-based benchmarks (24 outcome-only and 25 outcome-augmented) and 15 process-diagnostic benchmarks by system family. The matrix reports the proportions of Explicit, Implicit, and Not Evaluated labels for each PAPAV capability within each family.
\subsubsection{Overall PAPAV Coverage}
Because the corpus contains 10 MMEA, 23 MMA, and 31 RS benchmarks, comparisons across families are descriptive rather than estimates of the complete literature.

Act has the strongest explicit coverage: 10/10 (100\%) MMEA, 16/23 (70\%) MMA, and 29/31 (94\%) RS benchmarks. Perceive is Explicit in 7/10 (70\%) MMEA, 10/23 (43\%) MMA, and 10/31 (32\%) RS benchmarks. Explicit Perceive requires a dedicated understanding, grounding, or evidence acquisition task. Visual input alone is insufficient.

Plan shows the clearest difference. It is Explicit in 7/10 (70\%) MMEA benchmarks and Implicit in the other 3/10 (30\%). It is Implicit in 16/23 (70\%) MMA and 13/31 (42\%) RS benchmarks, with no Explicit cases in either family. The remaining 7/23 (30\%) MMA and 18/31 (58\%) RS benchmarks do not evaluate Plan. This pattern reflects protocol design, not the planning ability of each system family.

Only 5/64 (8\%) benchmarks explicitly evaluate Anticipate, and 3/64 (5\%) explicitly evaluate Verify. Explicit Evaluation of Anticipate requires a scored prediction of future states or action consequences. Verify requires assessment of the agent's success or failure judgments from feedback.

\subsubsection{Diagnosticity and Capability Bottlenecks}
Broad capability coverage does not guarantee useful failure diagnosis. A complex task may require perception, planning, and action but still report only a single overall success score. WorldModelBench~\citep{li2026worldmodelbench} and PBench~\citep{nvidia2025pbench} use focused protocols to isolate physical future prediction. HarnessEval-W~\citep{mirros2026harnessevalw} adds evidence-grounded diagnosis of predicted transitions and world persistence. Its evaluation traces explain scoring decisions, but do not expose the evaluated model's internal decision process or establish its Verify capability. AgentStudio's CriticBench~\citep{zheng2025agentstudio} directly evaluates task success detection. RoboCerebra~\citep{han2026robocerebra} records hierarchical decisions and intermediate checking signals. AgenticLab~\citep{guo2026planar} evaluates closed-loop replanning after observed execution deviations. These process measures provide more direct evidence and help distinguish poor planning, missed failures, and faulty execution.

Dream.exe~\citep{zhao2026dream} and RoboWM-Bench~\citep{jiang2026robowm} score useful physical predictions but stop before the agent compares alternatives, acts, and updates its forecast from feedback. Unit tests, simulator checkers, and safety monitors assess outcomes for the benchmark. Verify can be assessed through the agent's success or failure judgments. Further evaluation is needed to determine whether these judgments improve subsequent behavior. Explicit Plan similarly requires scored decomposition, subgoal order, constraints, coordination, or replanning. Controlled input interventions can reveal failures hidden by aggregate success rates. In LIBERO-Plus~\citep{fei2025libero}, instruction removal and target replacement tests probe whether successful execution depends on the specified goal. These comparisons provide evidence about grounding, although they do not fully separate perceptual and control errors.

\subsubsection{Convergence across Domains}
Interactive evaluation is becoming more similar across domains. VisualWebArena~\citep{koh2024visualwebarena} and OSWorld~2.0~\citep{yuan2026osworld2} evaluate persistent interaction with digital states. RoboCasa365~\citep{nasiriany2026robocasa365} introduces language and compositional tasks into robotics. RoboTwin~\citep{mu2025robotwin} uses digital twins to increase environment variation. MMEA benchmarks further combine these directions. EMBODIEDBENCH~\citep{yang2025embodiedbench} covers several embodied settings. PARTNR~\citep{chang2025partnr} evaluates collaborative planning under heterogeneous constraints. These benchmarks increasingly share persistent states, multi-step interaction, multimodal feedback, and coordinated action.

Overall, across the three domains, Act receives the strongest explicit coverage, while Anticipate and Verify remain less observable. Differences across domains mainly reflect benchmark design rather than system capability. Process measures provide clearer failure diagnosis than aggregate task scores. Benchmarks from different domains also show increasingly similar interaction and evaluation settings.

\subsection{Evaluation Metrics and Reproducibility}
This section examines evaluation metrics and reproducibility of the curated benchmarks. Evaluation metrics affect how well a benchmark can identify the source of failure. Reproducibility shows whether results can be repeated under the same experimental settings. Differences in tasks, interfaces, and environments make raw scores difficult to compare across benchmarks.

\subsubsection{Outcome- and Process-Based Evaluation}
We classified the benchmarks into three categories based on their evaluation metrics: outcome-only, outcome-augmented, and process-diagnostic. Outcome-only benchmarks report only terminal results, such as answer accuracy~\citep{mialon2024gaia}, task success~\citep{mees2022calvin,zhou2024webarena}, test pass rate~\citep{jimenez2024swe}, or artifact quality~\citep{lu2026webgen,luo2026gamecraft}. Outcome-augmented benchmarks also report progress, efficiency, cost, or constraint violations, but these measures do not link performance to a specific capability~\citep{anderson2018vision,khanna2024goat,shridhar2020alfred}. Process-diagnostic benchmarks include at least one metric, annotation, intermediate assessment, or failure category that links performance to a specific capability~\citep{xu2025crab,lu2026bench,guo2026planar}.

As shown in the left panel of Figure~\ref{fig:benchmark-analysis}, the corpus contains 24 outcome-only, 25 outcome-augmented, and 15 process-diagnostic benchmarks. Process diagnosis appears in 4/10~(40\%) MMEA, 4/23~(17.39\%) MMA, and 7/31~(22.58\%) RS benchmarks. Outcome-based metrics are easy to interpret but weak at locating PAPAV failures. Process-based metrics improve attribution but require finer annotations, trajectory logging, and evaluator design.

\subsubsection{Reproducibility}
Benchmark reproducibility depends on the stability of task data, execution environments, and evaluators. Static question answering and offline generation use fixed inputs and often deterministic scoring, making them easier to reproduce. Self-hosted websites, operating systems, and simulated environments offer more control but remain sensitive to browser, dependency, simulator, and model-interface versions. Real-robot evaluation introduces further variation from hardware, calibration, control frequency, scene configuration, and human intervention.

Procedural success checkers, environment snapshots, simulator reconstruction, and standardized hardware improve stability~\citep{zhou2024webarena,rawles2025androidworld,huang2026vla}. Distributed and reconstructed evaluations such as RoboArena~\citep{roboarena2025} and RobotArena$\infty$~\citep{jangir2025robotarena} still vary with task selection, execution conditions, and evaluator implementation. Benchmarks should report task instances, environment and evaluator versions, trajectories, model and prompt configurations, interfaces, budgets, random seeds, trial counts, and hardware settings. Human or model-based evaluation also requires documented criteria and consistency checks.

For this survey, the corpus was frozen at the stated cutoff date. Table~\ref{tab:benchmark-main-table} provides the retained entities, category assignments, and PAPAV labels needed to reproduce the capability coverage counts.

\subsection{Discussion and Future Benchmark Design}
A recurring limitation in this corpus is limited evaluation granularity. Multimodal, long-horizon, and interactive tasks increasingly couple several PAPAV capabilities across digital and embodied environments~\citep{zhou2024webarena,yuan2026osworld2,padmakumar2022teach,chang2025partnr}, yet aggregate outcomes rarely identify the source of failure. Future benchmarks should retain end-to-end performance while adding attributable signals such as grounding correctness, failure recognition, recovery effectiveness, controlled perturbations, and intermediate checkpoints. RoboCerebra~\citep{han2026robocerebra} and AgenticLab~\citep{guo2026planar} illustrate how checking and replanning signals can improve failure attribution within the PAPAV loop.

Future benchmarks should test how information passes between PAPAV capabilities and changes later decisions. They should record whether predictions affect action selection and whether execution feedback leads to revised plans. Verification requires particular care because an apparently completed action may not satisfy the task goal. A grasp, for example, may look correct while remaining unstable. REFLECT~\citep{liu2023reflect} uses observations of robot execution to explain failures and guide correction. Benchmarks should assess both the accuracy of these judgments and their effect on subsequent behavior.

The evidence available to the verifier is central to this evaluation. A dedicated verifier can specialize in detecting success and failure. A shared model can reuse the actor's representations and context, which may reduce computation. However, separate models may still miss the same failures if they use the same observations or similar training data. Benchmarks should therefore test whether additional evidence helps the verifier detect errors that the actor overlooks. Such evidence may come from another sensing modality, a physical check, a specialized success detector, or human review.

The actor and verifier should then be evaluated together. Benchmarks should report the fraction of failed executions judged successful and the fraction of successful executions judged failed. They should also identify failure types that both components miss. Detection delay, recovery success, and the extra actions or time needed for recovery show whether verification helps the task. Comparisons should control task conditions and report the resources available to each system. Reference judgments should be established independently of the agent's verdict. These measures assess both the reliability of verification and its contribution to recovery.

These evaluations should also include realistic deployment conditions. RoboArena~\citep{roboarena2025} and RobotArena$\infty$~\citep{jangir2025robotarena} illustrate how results can vary across sites and reconstructed environments. Benchmarks should combine controlled comparisons with physical trials or environmental perturbations. This would help determine whether verification and recovery remain effective when observations, hardware, or execution conditions change. The objective is to obtain reliable evidence about the capabilities and feedback processes required by each task.

%% file: tables/Benchmark_Table.tex
\begingroup
\scriptsize
\setlength{\tabcolsep}{1.5pt}
\setlength{\LTleft}{0pt}
\setlength{\LTright}{0pt}
\renewcommand{\arraystretch}{1.18}

% Make longtable caption use full text width
\setlength{\LTcapwidth}{\textwidth}

\begin{longtable}{
@{}
L{0.23\linewidth}
C{0.14\linewidth}
@{\hspace{6pt}}
L{0.24\linewidth}
!{\color{tabledivider}\vrule width 0.5pt}
*{5}{C{0.066\linewidth}}
@{}
}

\caption[Benchmark Overview and PAPAV Capability Coverage]{
\textbf{Overview and PAPAV coverage of 64 curated benchmarks (10 MMEA, 23 MMA, and 31 RS)}.
Per./Ant./Ver. denote Perceive/Anticipate/Verify, respectively.
ACL-F, CVPRW, and NeurIPS-DB denote Findings of ACL, CVPR Workshop, and NeurIPS Datasets and Benchmarks, respectively.
}
\label{tab:benchmark-main-table}\\

\toprule
\textbf{Benchmark} &
\textbf{Venue} &
\textbf{Task} &
\textbf{Per.} &
\textbf{Ant.} &
\textbf{Plan} &
\textbf{Act} &
\textbf{Ver.} \\
\midrule
\endfirsthead

\multicolumn{8}{r}{\scriptsize\itshape Continued} \\
\toprule
\textbf{Benchmark} &
\textbf{Venue} &
\textbf{Task} &
\textbf{Per.} &
\textbf{Ant.} &
\textbf{Plan} &
\textbf{Act} &
\textbf{Ver.} \\
\midrule
\endhead

\midrule
\multicolumn{8}{r}{\scriptsize\itshape Continued on next page} \\
\endfoot

\bottomrule
\multicolumn{8}{@{}l}{\scriptsize
\textbf{Evaluation:} \directcov\ Explicit \quad
\partialcov\ Implicit \quad
\nonecov\ Not Evaluated.} \\
\endlastfoot

% ============================================================
% Multimodal Embodied Agent
% ============================================================

\addlinespace[3pt]
\rowcolor{mmembodied}
\multicolumn{8}{@{}l}{{\small\bfseries Category: Multimodal Embodied Agent}} \\

\addlinespace[1pt]
\rowcolor{subcategory}
\multicolumn{8}{@{}l}{{\scriptsize\bfseries Subcategory: Simulation}} \\

TEACh~\citep{padmakumar2022teach} &
AAAI'22 &
Dialogue-guided tasks &
\partialcov & \partialcov & \partialcov & \directcov & \partialcov \\

RoCoBench~\citep{mandi2024roco} &
ICRA'24 &
Collaborative coordination &
\nonecov & \partialcov & \directcov & \directcov & \partialcov \\

PARTNR~\citep{chang2025partnr} &
ICLR'25 &
Collaborative planning &
\directcov & \partialcov & \directcov & \directcov & \partialcov \\

EMBODIEDBENCH~\citep{yang2025embodiedbench} &
ICML'25 &
Embodied tasks &
\directcov & \partialcov & \directcov & \directcov & \partialcov \\

RoboCerebra~\citep{han2026robocerebra} &
NeurIPS'25 &
Long-horizon planning &
\partialcov & \partialcov & \directcov & \directcov & \directcov \\

Habitat-MAS~\citep{chen2025emos} &
ICLR'25 &
Multi-robot allocation &
\directcov & \partialcov & \directcov & \directcov & \partialcov \\

IS-Bench~\citep{lu2026bench} &
AAAI'26 &
Hazard mitigation &
\directcov & \partialcov & \partialcov & \directcov & \nonecov \\

ESI-Bench~\citep{hong2026esi} &
arXiv'26 &
Spatial intelligence &
\directcov & \partialcov & \directcov & \directcov & \nonecov \\

\addlinespace[1pt]
\rowcolor{subcategory}
\multicolumn{8}{@{}l}{{\scriptsize\bfseries Subcategory: Real}} \\

AgenticLab~\citep{guo2026planar} &
arXiv'26 &
Verified replanning &
\directcov & \partialcov & \directcov & \directcov & \directcov \\

\addlinespace[1pt]
\rowcolor{subcategory}
\multicolumn{8}{@{}l}{{\scriptsize\bfseries Subcategory: Hybrid}} \\

CaP-Bench~\citep{fu2026cap} &
arXiv'26 &
Robot code refinement &
\directcov & \partialcov & \partialcov & \directcov & \partialcov \\

% ============================================================
% Multimodal Agent
% ============================================================

\addlinespace[3pt]
\midrule
\rowcolor{mmagent}
\multicolumn{8}{@{}l}{{\small\bfseries Category: Multimodal Agent}} \\

\addlinespace[1pt]
\rowcolor{subcategory}
\multicolumn{8}{@{}l}{{\scriptsize\bfseries Subcategory: Understanding}} \\

MMMU-Pro~\citep{yue2025mmmu} &
ACL'25 &
Multimodal QA &
\directcov & \nonecov & \nonecov & \nonecov & \nonecov \\

VSI-Bench~\citep{yang2025thinkinginspace} &
CVPR'25 &
Spatial reasoning &
\directcov & \nonecov & \nonecov & \nonecov & \nonecov \\

Video-MME-v2~\citep{fu2026video} &
arXiv'26 &
Video QA &
\directcov & \nonecov & \nonecov & \nonecov & \nonecov \\

\addlinespace[1pt]
\rowcolor{subcategory}
\multicolumn{8}{@{}l}{{\scriptsize\bfseries Subcategory: Interaction}} \\

WebArena~\citep{zhou2024webarena} &
ICLR'24 &
Web workflows &
\nonecov & \partialcov & \partialcov & \directcov & \partialcov \\

GAIA~\citep{mialon2024gaia} &
ICLR'24 &
General AI Tasks &
\partialcov & \nonecov & \partialcov & \partialcov & \nonecov \\

VisualWebArena~\citep{koh2024visualwebarena} &
ACL'24 &
Visual web tasks &
\directcov & \partialcov & \partialcov & \directcov & \partialcov \\

WebVoyager~\citep{he2024webvoyager} &
ACL'24 &
Live web tasks &
\partialcov & \partialcov & \partialcov & \directcov & \partialcov \\

AndroidWorld~\citep{rawles2025androidworld} &
ICLR'25 &
Android tasks &
\partialcov & \partialcov & \partialcov & \directcov & \partialcov \\

AgentStudio~\citep{zheng2025agentstudio} &
ICLR'25 &
Virtual-agent tasks &
\directcov & \partialcov & \partialcov & \directcov & \directcov \\

MMInA~\citep{tian2025mmina} &
ACL-F'25 &
Multimodal web tasks &
\directcov & \partialcov & \partialcov & \directcov & \nonecov \\

CRAB~\citep{xu2025crab} &
ACL-F'25 &
Cross-device workflows &
\partialcov & \partialcov & \partialcov & \directcov & \partialcov \\

iVISPAR~\citep{mayer2025ivispar} &
EMNLP'25 &
Sliding-tile puzzles &
\directcov & \partialcov & \nonecov & \directcov & \nonecov \\

AgentVista~\citep{su2026agentvista} &
arXiv'26 &
Visual agent tasks &
\directcov & \nonecov & \partialcov & \directcov & \nonecov \\

GameWorld~\citep{ouyang2026gameworld} &
arXiv'26 &
Browser-game tasks &
\partialcov & \partialcov & \partialcov & \directcov & \nonecov \\

MMSearch-Plus~\citep{tao2026mmsearch} &
ICLR'26 &
Multimodal retrieval &
\directcov & \nonecov & \partialcov & \directcov & \nonecov \\

OmniGAIA~\citep{li2026omnigaia} &
arXiv'26 &
Omni-modal QA &
\directcov & \nonecov & \partialcov & \directcov & \nonecov \\

OSWorld2.0~\citep{yuan2026osworld2} &
arXiv'26 &
Computer-use Task &
\partialcov & \partialcov & \partialcov & \directcov & \partialcov \\

\addlinespace[1pt]
\rowcolor{subcategory}
\multicolumn{8}{@{}l}{{\scriptsize\bfseries Subcategory: Generation}} \\

SWE-bench~\citep{jimenez2024swe} &
ICLR'24 &
Repository repair &
\partialcov & \nonecov & \partialcov & \directcov & \nonecov \\

WebGen-Bench~\citep{lu2026webgen} &
NeurIPS-DB'25 &
Interactive website generation &
\partialcov & \nonecov & \partialcov & \directcov & \nonecov \\

PBench~\citep{nvidia2025pbench} &
NVIDIA'25 &
Physical future prediction &
\partialcov & \directcov & \nonecov & \nonecov & \nonecov \\

GameCraft-Bench~\citep{luo2026gamecraft} &
arXiv'26 &
Playable game generation &
\partialcov & \nonecov & \partialcov & \directcov & \nonecov \\

WorldModelBench~\citep{li2026worldmodelbench} &
NeurIPS-DB'25 &
Future video generation &
\partialcov & \directcov & \nonecov & \nonecov & \nonecov \\

HarnessEval-W~\citep{mirros2026harnessevalw} &
arXiv'26 &
Interactive world prediction &
\partialcov & \directcov & \nonecov & \nonecov & \nonecov \\

% ============================================================
% Robotic System
% ============================================================

\addlinespace[3pt]
\midrule
\rowcolor{robotics}
\multicolumn{8}{@{}l}{{\small\bfseries Category: Robotic System}} \\

\addlinespace[1pt]
\rowcolor{subcategory}
\multicolumn{8}{@{}l}{{\scriptsize\bfseries Subcategory: Simulation}} \\

Room-to-Room (R2R)~\citep{anderson2018vision} &
CVPR'18 &
Instruction navigation &
\directcov & \partialcov & \nonecov & \directcov & \partialcov \\

ALFRED~\citep{shridhar2020alfred} &
CVPR'20 &
Household tasks &
\directcov & \partialcov & \partialcov & \directcov & \partialcov \\

RLBench~\citep{james2020rlbench} &
RA-L'20 &
Manipulation learning &
\partialcov & \partialcov & \nonecov & \directcov & \nonecov \\

CALVIN~\citep{mees2022calvin} &
RA-L'22 &
Instruction sequences &
\directcov & \partialcov & \nonecov & \directcov & \nonecov \\

BEHAVIOR-1K~\citep{li2023behavior} &
CoRL'22 &
Household activities &
\partialcov & \partialcov & \partialcov & \directcov & \partialcov \\

ManiSkill2~\citep{gu2023maniskill2} &
ICLR'23 &
Generalizable manipulation &
\partialcov & \partialcov & \nonecov & \directcov & \nonecov \\

LIBERO~\citep{liu2023libero} &
NeurIPS-DB'23 &
Lifelong manipulation &
\partialcov & \partialcov & \partialcov & \directcov & \nonecov \\

Safety-Gymnasium~\citep{ji2023safety} &
NeurIPS'23 &
Safe control &
\partialcov & \partialcov & \nonecov & \directcov & \nonecov \\

GOAT-Bench~\citep{khanna2024goat} &
CVPR'24 &
Multimodal navigation &
\directcov & \partialcov & \nonecov & \directcov & \partialcov \\

RoboSuite~\citep{zhu2020robosuite} &
arXiv'25 &
Policy training &
\partialcov & \nonecov & \nonecov & \directcov & \nonecov \\

VLABench~\citep{zhang2025vlabench} &
ICCV'25 &
Long-horizon manipulation &
\directcov & \partialcov & \partialcov & \directcov & \partialcov \\

Meta-World +~\citep{mclean2026meta} &
NeurIPS-DB'25 &
Multi-skill learning &
\nonecov & \nonecov & \nonecov & \directcov & \nonecov \\

LIBERO-Plus~\citep{fei2025libero} &
CVPR'26 &
Robust manipulation &
\directcov & \partialcov & \partialcov & \directcov & \nonecov \\

LIBERO-PRO~\citep{zhou2025libero} &
arXiv'26 &
Robust manipulation &
\directcov & \partialcov & \partialcov & \directcov & \nonecov \\

SafeManip~\citep{huang2026safemanip} &
arXiv'26 &
Safe manipulation &
\partialcov & \partialcov & \partialcov & \directcov & \nonecov \\

Dream.exe~\citep{zhao2026dream} &
arXiv'26 &
Future video execution &
\partialcov & \directcov & \nonecov & \partialcov & \nonecov \\

RoboCasa365~\citep{nasiriany2026robocasa365} &
ICLR'26 &
Kitchen activities &
\partialcov & \partialcov & \partialcov & \directcov & \nonecov \\

RoboLab~\citep{yang2026robolab} &
RSS'26 &
Robust manipulation &
\directcov & \partialcov & \partialcov & \directcov & \nonecov \\

\addlinespace[1pt]
\rowcolor{subcategory}
\multicolumn{8}{@{}l}{{\scriptsize\bfseries Subcategory: Real}} \\

FurnitureBench~\citep{heo2023furniturebench} &
RSS'23 &
Furniture assembly &
\partialcov & \partialcov & \partialcov & \directcov & \partialcov \\

FMB~\citep{luo2025fmb} &
IJRR'24 &
Skill composition &
\partialcov & \partialcov & \partialcov & \directcov & \nonecov \\

RoboArena~\citep{roboarena2025} &
CoRL'25 &
Policy comparison &
\partialcov & \nonecov & \nonecov & \directcov & \nonecov \\

BEHAVIOR Robot Suite~\citep{jiang2025behavior} &
CoRL'25 &
Household manipulation &
\partialcov & \partialcov & \partialcov & \directcov & \partialcov \\

RoboChallenge~\citep{robochallenge2025} &
Tech. Rep.'25 &
Policy evaluation &
\partialcov & \nonecov & \nonecov & \directcov & \nonecov \\

VLA-REPLICA~\citep{huang2026vla} &
arXiv'26 &
Reproducible manipulation &
\partialcov & \nonecov & \nonecov & \directcov & \nonecov \\

\addlinespace[1pt]
\rowcolor{subcategory}
\multicolumn{8}{@{}l}{{\scriptsize\bfseries Subcategory: Hybrid}} \\

HomeRobot~\citep{yenamandra2023homerobot} &
CoRL'23 &
Mobile manipulation &
\directcov & \partialcov & \partialcov & \directcov & \partialcov \\

RoboTwin~\citep{mu2025robotwin} &
CVPR'25 &
Bimanual transfer &
\partialcov & \nonecov & \nonecov & \directcov & \nonecov \\

RobotArena$\infty$~\citep{jangir2025robotarena} &
ICLR'26 &
Reconstructed evaluation &
\directcov & \nonecov & \nonecov & \directcov & \nonecov \\

RoboTwin 2.0~\citep{chen2025robotwin} &
ICML'26 &
Bimanual policy generation &
\partialcov & \nonecov & \nonecov & \directcov & \nonecov \\

RoboWM-Bench~\citep{jiang2026robowm} &
CVPRW'26 &
Future manipulation &
\partialcov & \directcov & \nonecov & \partialcov & \nonecov \\

NIABench~\citep{zhang2026assistance} &
IROS'26 &
Non-intrusive assistance &
\nonecov & \partialcov & \nonecov & \directcov & \nonecov \\

RoboDojo~\citep{robodojo2026} &
arXiv'26 &
Sim-to-real evaluation &
\partialcov & \nonecov & \nonecov & \directcov & \nonecov \\

\end{longtable}
\endgroup

%% file: sections/5-challenges.tex
\section{Open Challenges}
\label{sec:challenges}

The following challenges are not intended to cover every open problem in embodied AI. They instead arise from a common difficulty: PAPAV capabilities must remain coordinated while the agent interacts with a world that is only partially observed, continuously changing, difficult to simulate, and often hard to verify. Under these conditions, it is not enough for each capability to improve in isolation. The agent must decide what information to retain, which future consequences to predict, when to revise a plan, how to act under uncertainty, and how verification can improve later behavior. We organize these five challenges in PAPAV order.

\subsection{\textcolor{perceive}{OC1}. Persistent Physical Belief: What Should the Agent Remember?}
\label{sec:challenge-memory}

A robot never has the physical world fully available in its current
observation. Objects move out of view, properties become hidden by
occlusion, and the effects of earlier actions can remain important long after
the evidence has disappeared. Classical state estimation and mapping
already provide mechanisms for carrying information forward in time
\citep{campos2021orb}; more recent open-vocabulary maps and scene graphs extend
this idea toward persistent semantic entities and relations
\citep{gu2024conceptgraphs,maggio2024clio}. Yet persistence alone is not enough.
Unlike a static log, a physical belief must change when the world changes.

This creates a tension that is easy to overlook. Forgetting too aggressively
causes the agent to repeatedly rediscover state it has already observed, while
retaining everything allows stale or contradictory information to accumulate.
Structured maps make beliefs explicit and editable; latent recurrent states can
compress long histories; episodic retrieval can recover fine-grained evidence
only when it becomes relevant. None of these properties alone determines what
the agent should continue to believe.

\begin{center}
\textit{\textbf{Core question:} What should an embodied agent retain, revise, or forget as the world changes?}
\end{center}

The central issue is therefore not memory capacity, but the quality of the
belief that memory supports. A useful memory should preserve information that
is temporarily unobservable, revise it when new evidence disagrees, and make
past experience available when it can change the next decision. This suggests
that future systems may need to combine persistent structured state with
compressed or episodic representations rather than choosing a single memory
form.

Correspondingly, evaluation should ask whether memory remains useful under
occlusion and change: can the agent recover hidden-but-relevant state, correct
stale beliefs, resolve contradictions, and make better downstream decisions
without memory and retrieval cost growing without bound?

\subsection{\textcolor{anticipate}{OC2}. Decision-Relevant Prediction: What Is Worth Predicting?}
\label{sec:challenge-world-model}

Embodied agents can anticipate the future at very different levels of detail.
Visual foresight predicts future observations under candidate actions
\citep{finn2017deep,ebert2017self}; latent world models instead compress the
dynamics into an internal state and reason through imagined trajectories
\citep{hafner2025dreamerv3,pmlr-v205-wu23c}. Recent robotic world models further
explore learned predictive representations for manipulation and robot
imagination \citep{robodreamer2024,irasim2024}. These approaches all support
anticipation, but they do not agree on what must actually be predicted.

A high-fidelity future image may contain nearly every visible consequence of an
action, yet much of that detail may be irrelevant to the choice at hand.
Conversely, a compact latent prediction can be efficient while hiding the very
factor that determines whether an action is safe or feasible. For some
decisions, predicting an event such as collision likelihood, contact,
reachability, progress, or risk may be more useful than reconstructing the
future scene itself \citep{kahn2018composable}.

\begin{center}
\textit{\textbf{Core question:} What future information must an embodied agent
predict to make a good decision?}
\end{center}

This suggests a different objective for world modeling. Rather than asking for the most realistic or complete forecast, the agent should predict enough of the future to preserve the distinctions that matter for action. What counts as sufficient prediction may therefore depend on both the task and the decision the agent is trying to make. The desired representation may change accordingly: geometric consequences may dominate navigation, contact and stability may matter for manipulation, and social responses or intent may become more important in human--robot interaction.

A practical way to test such sufficiency is to remove predictive detail and ask
whether the resulting policy changes. If a simpler forecast preserves action
ranking, safety judgments, and task outcomes, the omitted detail was not
necessary for that decision. The open problem is to identify such
decision-relevant representations without making them so narrow that they fail
when the environment changes.

\subsection{\textcolor{plan}{OC3}. Adaptive Commitment: When Should the Agent Replan?}
\label{sec:challenge-granularity}

The granularity of action has become an increasingly important design choice.
Policies such as Action Chunking with Transformers predict temporally coherent
action sequences rather than isolated controls \citep{zhao2023act}. This can
reduce repeated high-level inference and produce smoother execution. At the same
time, longer chunks create a period during which the world may change while the
agent is still committed to an earlier decision. Real-Time Chunking makes this
tension explicit by asking how large policies can remain reactive when model
inference is slower than physical execution \citep{black2025rtc}.

Neither extreme is satisfactory. Replanning after every small motion is
expensive and can make behavior hesitant; committing to a long action sequence
can make the agent slow to react to contact, disturbances, or an incorrect
belief. Closed-loop systems such as Inner Monologue already show why new scene
and success information should feed back into later decisions
\citep{huang2022inner}. What remains unclear is how frequently that loop should
be reopened.

\begin{center}
\textit{\textbf{Core question:} How long should an agent continue
executing before observing and replanning?}
\end{center}

The appropriate horizon is unlikely to be fixed. Predictable free-space motion
may tolerate longer commitment, whereas contact-rich interaction, unexpected
motion, high uncertainty, or an irreversible action may justify checking again
much sooner. In this view, action granularity becomes a control variable of the
agent itself: the system adapts not only \emph{what} it does, but also
\emph{when} it should think again.

A convincing solution should therefore gain reactivity where it matters without
paying the cost of constant replanning everywhere. Success rate alone will not
capture this trade-off; reaction delay, wasted actions, intervention frequency,
and inference cost are also important for determining whether adaptive
commitment actually improves the closed loop.

\subsection{\textcolor{act}{OC4}. Uncertainty-Conditioned Control: Act, Observe, or Ask?}
\label{sec:challenge-uncertainty}

Uncertainty appears throughout the embodied loop. An object may be partially occluded, an instruction may admit several interpretations, a scene may differ from the training distribution, or the outcome of an action may simply be hard to predict. In each case, uncertainty can affect not only what the agent believes, but also which action it should take and how cautiously it should proceed. Recent work has begun to make this uncertainty actionable: calibrated planners can ask for help \citep{ren2023knowno}, active-perception systems can gather additional observations \citep{ghasemi2020task,activevla2026}, and runtime monitors can use uncertainty to detect situations in which the current policy may be unreliable \citep{xu2025can,introspective2024}.

These mechanisms point to a control problem. More information is not always worth acquiring, and uncertainty does not always justify stopping. A slightly uncertain but reversible action may be worth attempting; the same uncertainty may be unacceptable before an irreversible or safety-critical action. Likewise, asking for help can prevent failure but defeats autonomy if used whenever confidence drops.

\begin{center}
\textit{\textbf{Core question:} When uncertain, should an agent act,
gather more information, or ask for help?}
\end{center}

The choice depends on more than calibrated confidence. The agent must also
consider how costly another observation would be, whether it is likely to
resolve the ambiguity, how reversible the proposed action is, and what happens
if the decision is wrong. Uncertainty estimation therefore becomes useful only
when it changes the control strategy of the loop.

This reframes the goal from ``knowing when the model is uncertain'' to choosing
an appropriate response to that uncertainty. A capable agent should proceed
when autonomous action is justified, seek information when it is likely to
change the decision, and defer when the potential consequence of a mistake
outweighs the cost of asking. The challenge is to achieve this balance without
producing either reckless autonomy or a system that continually re-observes and
escalates.

\subsection{\textcolor{verify}{OC5}. Learning from Verification: How Can the Agent Improve?}\label{sec:challenge-verification}
Verification judges whether an action achieved its intended outcome. Its role in an improvement loop extends to guiding changes in later behavior. REFLECT~\citep{liu2023reflect} uses robot experience to explain failures and support correction. Reflexion~\citep{shinn2023reflexion} stores reflections on task feedback and uses them in later trials. These approaches motivate a broader question: how can feedback produce improvements that remain useful beyond the current attempt and across related tasks?

A central difficulty is deciding what should change. A failed manipulation may result from an incorrect object estimate, a poor prediction, an unsuitable plan, or an execution error. The same failure signal can therefore call for different updates across PAPAV. Revising the wrong component may leave the cause untouched or introduce new errors. The agent must connect the observed outcome to the assumptions and decisions that produced it, using evidence from execution.

\begin{center}
\textit{\textbf{Core question:} How can verification guide reliable improvement across the PAPAV loop?}
\end{center}

An effective improvement loop must connect a judgment to a specific revision and test that revision through further interaction before retaining it for reuse. Feedback may correct a belief, update a prediction, revise a planning rule, or refine an execution strategy. Such changes can occur through memory and decision updates without changing model parameters. The difficult step is determining whether a revision addresses the cause of failure and when it remains applicable.

A successful retry alone does not establish lasting improvement. A correction may work in one scene but fail when objects, goals, or conditions change. Repeated feedback may also reinforce an incorrect explanation and spread its effects across the loop. Agents therefore need ways to retain useful lessons, limit their application to relevant situations, and reverse changes when new evidence contradicts them. The open challenge is to accumulate useful improvements while preventing errors in verification from accumulating through repeated updates and undermining previously acquired skills.

%% file: sections/6-conclusions.tex
\section{Conclusion}
\label{sec:conclusion}
This paper has asked what changes when a multimodal agent moves from computer-use to robot-use. To answer it, we introduced \papav{}, a unified capability-centric perspective that compares multimodal agents, robotic systems, and MMEAs through five functions in a shared task loop. Because each capability is defined by its contribution to the loop rather than by architecture, the digital--physical transition becomes directly comparable across research traditions.

The comparison reveals a consistent shift. Embodiment preserves the task loop but subjects it to partial observability, limited simulatability, irreversibility, real-time coupling, and unverifiable outcomes. Choices that structured interfaces or system design settle in advance for computer-use agents remain open once the same agent uses a robot, so more decisions move into the interaction itself. Reliability therefore depends on how well an agent maintains its beliefs, qualifies its predictions, authorizes its commitments, keeps execution visible, and checks outcomes as physical consequences unfold.

The benchmark analysis points to a broader lesson: end-to-end success is most informative when the underlying capabilities are also observable. Explicit evaluation currently concentrates on Act, while Anticipate and Verify remain largely hidden behind task success. Combining capability-level tests with end-to-end tasks would make failures easier to locate and reported progress easier to interpret. The open challenges we identify, from what an agent should remember to how it should act under uncertainty, follow from the same constraints and mark where the loop must be coordinated rather than merely assembled. We hope PAPAV offers a common basis for designing reliable physical agents and for measuring progress across the full interaction loop.